\documentclass{article}
\usepackage[T1]{fontenc}
\usepackage[utf8]{inputenc}
\usepackage{microtype,graphicx,booktabs}
\usepackage[hidelinks]{hyperref}

\usepackage[preprint]{icml2027}
\usepackage{amsmath,amssymb,amsthm,mathtools,array}
\usepackage{algorithm,algorithmic}
\newtheorem{theorem}{Theorem}[section]

\newtheorem{proposition}[theorem]{Proposition}
\newtheorem{corollary}[theorem]{Corollary}
\newcommand{\E}{\mathbb E}
\newcommand{\R}{\mathbb R}
\newcommand{\norm}[1]{\left\lVert#1\right\rVert}
\newcommand{\tr}{\operatorname{tr}}

\newcommand{\PSD}{\succeq 0}
\newcommand{\loss}{\ell_{.95}}
\newcommand{\Normal}{\mathcal N}
\newcommand{\ResultsAbstractSentence}{Evaluation uses twenty new reference contexts per dataset at two task sizes, giving 400 conditions across ten benchmark populations. The mean dataset complete-loss ratios are 0.9977 to the degree-two kernel, 0.9961 to matched complete-MC shrinkage, and 0.9911 to the retrospective minimum over 174 public configurations. The last ratio has a paired reference-bootstrap interval of [0.9898,0.9927]. On 40 profiles at coefficient tolerance $10^{-7}$, Gamma quadrature has a geometric mean cold-time ratio of .0646 to strengthened Arb/Petras integration.}
\newcommand{\EvaluationProtocolSentence}{Method development uses four reference contexts and four Monte Carlo streams per condition. After fixing the prediction rule, evaluation uses twenty new reference contexts per dataset and 2048 evaluation tasks per condition.}

\icmltitlerunning{Calibrated Functional Prediction from Nonlinear Observations}
\hypersetup{pdfauthor={Peida Li},pdftitle={Calibrated Functional Prediction from Overlapping Nonlinear Observations}}
\renewcommand{\today}{November 16, 2026}
\begin{document}
\twocolumn[
\icmltitle{Calibrated Functional Prediction from\\Overlapping Nonlinear Observations}
\begin{icmlauthorlist}
\icmlauthor{Peida Li}{pku}
\end{icmlauthorlist}
\icmlaffiliation{pku}{School of Basic Medical Sciences, Peking University}
\icmlcorrespondingauthor{Peida Li}{2410301207@stu.pku.edu.cn}
\icmlkeywords{aggregate observations, polynomial chaos, kernel learning, conformal prediction, validated numerics}
\vskip 0.3in
]
\hypersetup{pdftitle={Calibrated Functional Prediction from Overlapping Nonlinear Observations}}
\printAffiliationsAndNotice{}

\begin{abstract}
We study prediction of nonlinear audit functionals from scalar observations of a shared hidden table state.
For Schur-rational operators, local Gaussian projections yield a common sparse kernel for overlapping tasks, repeated row identities, and cross-row dependence.
Shared-state residual samples extend this kernel through a positive-semidefinite feature representation.
Positive matrix integration and sample-value enclosures connect the original inputs to the actual saved hybrid prediction.
A sorted calibration-role perturbation bound then certifies its complete finite-pool pinball risk.
Independent complete-loss selection chooses the ridge, score scale, and residual weight, while task-rank calibration supplies an upper endpoint.
\ResultsAbstractSentence
\end{abstract}

\section{Introduction}

A training table can support many related audit questions.
An audit may select a group of records and report how much a fixed predictor would lose against retraining after completing one feature.
Different audits reuse some of the same records and evaluate different nonlinear functions of the same hidden values.
Observed audit outcomes therefore constrain a shared state.
Predicting the outcome of another audit can use both its known operator and these shared constraints.

Aggregate-observation learning and distribution regression provide established ways to learn from grouped supervision.
Gaussian-process models allow overlapping bags, and polynomial-chaos filtering provides covariance-based updates from nonlinear observations.
These tools motivate an operator-aware computational problem.
Each audit can involve a small subset of a much larger hidden state, while a global degree-two expansion contains a quadratic number of coordinates.
Evaluating each pair of nonlinear operators separately also repeats integration work.
Accurate deployment additionally requires signed coefficient errors, omitted functional variation, and kernel-solve errors to be controlled together \citep{law2018,szabo2016,ernst2015,vondrejc2018}.

We compute each operator's local projection once, then lift it isometrically to a common Gaussian space.
Sparse row identities represent private coordinates, and a compact row-feature map represents cross-row dependence.
Their first- and second-order inner products give a positive-semidefinite task kernel.
For the rational operators considered here, a single positive matrix integral contains the mean and every required signed coefficient.
This integral admits matrix lower and upper bounds from positive Gauss--Radau rules.
The same tilted Gaussian calculation certifies a second moment and the higher-order residual energy.
We complete the task covariance with centered residual features evaluated on common states and select a finite shrinkage weight using independent calibrated loss.

Kernel ridge regression implements the regularized linear-Bayes update, and task-level rank calibration supplies the upper endpoint \citep{ernst2015,romano2019}.
Our method supplies an original-input construction and certificate for the structured functional kernel and its residual extension, including bounds for the saved prediction and complete calibrated loss.
Completion-relative ridge regret supplies a concrete operator class and an audit application.
\citet{li2026} establishes deterministic certificates for completion-relative retraining regret over continuous interval tables, with original-row residual conditions that yield uniform dimension reduction.
We use its regret formulation and Schur representation to study known task functionals evaluated on one shared hidden state.
Our contributions concern their common projection and residual kernel, validated numerical computation, and task-level calibrated prediction.

\begin{itemize}
\item We construct a common functional kernel from local projections and shared residual samples, with explicit treatment of repeated identities and cross-row Gaussian dependence.
\item We connect original-input matrix enclosures and finite-state function evaluation to the saved hybrid prediction and its complete calibrated-risk certificate.
\item We measure the predictive contribution of residual covariance, its computation cost, and learning gains from additional observed source functions on simulated and measured reference tables.
\end{itemize}

\section{Shared nonlinear task observations}
\label{sec:problem}

Let $v=(v_1,\ldots,v_D)$ denote the fixed hidden feature values of a reference table.
A task $i$ has a known selection map $P_i$ and a known scalar operator $h_i$.
Its observed outcome is $Y_i=h_i(P_i v)$.
We observe $n$ source outcomes and predict outcomes of new tasks drawn from the same task mechanism.
Observed row covariates, responses, reference predictors, and operator parameters are available task inputs.
Conditional on this observed reference context, the fixed hidden row state generates all scalar task outcomes.
The Gaussian working law below constructs their operator covariance approximation.

For the working distribution, write
\begin{equation}
v_j=m_j+\sigma_j\xi_j,\quad
\xi_j=\sqrt{1-\eta}\,\epsilon_j+\sqrt\eta\,\phi_j^\top g,
\label{eq:rowlaw}
\end{equation}
where $\epsilon_j$ and the coordinates of $g\in\R^d$ are independent standard Gaussians.
The row feature $\phi_j$ is normalized, $\sigma_j>0$, and $0\leq\eta<1$.
Reference-data fits supply $m_j$ and $\sigma_j$ using the estimators and fitting roles in Appendix~\ref{app:workingquantities}.
The experiments fix $\eta=1/2$.
All occurrences of row $j$ share the same $\xi_j$.
Repeated positions within a task are merged before local whitening.

We study nonnegative task functions with the form
\begin{equation}
\begin{aligned}
H_i(x)&=P_i(x)+R_i(x)^2/S_i(x),\\
P_i(x)&\geq0,\qquad S_i(x)\geq\delta_i>0,
\end{aligned}
\label{eq:rational}
\end{equation}
where $P_i,R_i,S_i$ have degree at most two and $x\sim\Normal(0,I_{q_i})$.
The ridge-regret application has an observed positive Schur floor inherited from its ridge parameter.
For a completed design $X$, weights $W$, and fixed candidate $\beta$, the operator is
\begin{align}
H(X)&=J_X(\beta)-\min_\theta J_X(\theta),\nonumber\\
J_X(\theta)&=(X\theta-y)^\top W(X\theta-y)+\lambda_0\norm{\theta}^2.
\label{eq:regret}
\end{align}
Original-input elimination gives Equation~\eqref{eq:rational}.
We use the established Schur algebra and retain all model columns \citep{li2026}.

\section{Local projections and a shared task kernel}
\label{sec:kernel}

For each task, compute the complete-function degree-two projection
\begin{equation}
H_i=\mu_i+a_i^\top x_i+\tfrac{1}{2}(x_i^\top B_i x_i-\tr B_i)+r_i.
\label{eq:projection}
\end{equation}
Let $F_i$ denote the centered polynomial term.
The residual $r_i$ is orthogonal to constants and to all first- and second-order polynomials in the common Gaussian space.

\begin{theorem}[Local-to-global functional kernel]
\label{thm:kernel}
Let $Z$ collect the common primitive coordinates.
For task $i$, write $x_i=A_iZ$ with $A_iA_i^\top=I$ and put $C_{ij}=A_iA_j^\top$.
Then
\begin{equation}
K_{ij}=a_i^\top C_{ij}a_j+\tfrac{1}{2}\tr(B_iC_{ij}B_jC_{ij}^\top)
\label{eq:kernel}
\end{equation}
is the Gram kernel $\E[F_iF_j]$.
The lift $(a_i,B_i)\mapsto(A_i^\top a_i,A_i^\top B_iA_i)$ preserves the norm $\norm{a_i}^2+\norm{B_i}_F^2/2$.
Moreover, $\operatorname{Cov}(H)=K+K_r$ with $K_r\PSD$ and
$|(K_r)_{ij}|\leq\rho_i\rho_j$, where $\rho_i=\norm{r_i}_{L^2}$.
\end{theorem}

For Equation~\eqref{eq:rowlaw}, the local covariance is
$R_{ii}=(1-\eta)I+\eta\Phi_i\Phi_i^\top$.
If $L_iL_i^\top=R_{ii}$, then $C_{ij}=L_i^{-1}R_{ij}L_j^{-\top}$.
The covariance cross block includes both matching row identities and the shared feature component.

The private row coordinates and the $d$ shared coordinates give a sparse representation with $O(q_i^2+q_id+d^2)$ coefficients per task.
Local integration has dimension $q_i$, and inner products of the lifted coefficients assemble the common kernel.

Given source kernel $K$, query cross kernel $k_*$, mean $\mu$, and $c>0$, the predictor is
\begin{equation}
\widehat H_* = \mu_*+\frac{k_*^\top}{c}
\left(\frac Kc+\lambda I\right)^{-1}(Y-\mu).
\label{eq:prediction}
\end{equation}
We set $c$ to the source mean kernel diagonal, with a fixed positive numerical floor.
The dimensionless ridge $\lambda$ corresponds to the original-unit penalty $\tau=c\lambda$.
Equation~\eqref{eq:prediction} applies a regularized linear-Bayes update using the means and degree-two covariance projections of the task operators \citep{ernst2015}.
The observed source outcomes are values of the complete operators.
\subsection{Residual covariance from shared states}
\label{sec:residualprediction}

The orthogonal decomposition in Theorem~\ref{thm:kernel} also supplies a common feature representation for the remaining covariance.
For $M\geq2$ shared states, put
\begin{equation}
u_i=\frac{(r_i(Z^{(b)})-\bar r_i)_{b=1}^M}{\sqrt{M-1}},\qquad
\widehat C_r=(u_i^\top u_j)_{ij}.
\end{equation}
We use the residual covariance kernel
\begin{equation}
K_\omega=K_2+\omega\widehat C_r,
\qquad \omega\in\{0,1/2,1\}.
\label{eq:residualkernel}
\end{equation}

\begin{proposition}[Finite-state task kernel]
\label{prop:residualkernel}
For every fixed set of states and $\omega\geq0$, $K_\omega$ is the Gram kernel of $(g_i,\sqrt\omega u_i)$, where $g_i$ is the common degree-two coefficient vector.
For a fixed $\omega$ and independent Gaussian states,
$\E K_\omega=K_2+\omega K_r$.
When $\omega=1$, it is unbiased and consistent for $\operatorname{Cov}(H)$.
Its covariance mean-square error depends on the residual fourth moment through the centered sample-covariance formula in Appendix~\ref{app:residualcompletion}.
\end{proposition}

The primitive state is shared across all tasks, and repeated row identities use the same coordinate.
Adding queries preserves every existing kernel block.
Polynomial-chaos control variates and multifidelity covariance methods provide the statistical foundations for analytic and sample-based combinations \citep{fox2021,yang2023,maurais2023}.
Here the known local operators provide the common low-order coefficients, while the residual samples provide query-compatible additional features.

\paragraph{When residual information changes the prediction.}
Let $A=K_{2,SS}+\tau I$, $R=K_{r,SS}$, $k=K_{2,S*}$, $t=K_{r,S*}$, and $y=Y_S-\mu_S$, with $\tau>0$.
For the population covariance, the complete and degree-two kernel predictors satisfy
\begin{equation}
p_H-p_2=(t-RA^{-1}k)^\top(A+R)^{-1}y.
\label{eq:residualdirection}
\end{equation}
The residual source--query direction $t-RA^{-1}k$ and its alignment with the updated source weights determine the change in prediction.
This explains why residual energy and predictive value can vary separately across task populations.

Finite-state estimation also introduces a bias--variance trade-off.
For a fixed weight independent of the Monte Carlo states, set $A_r=\|K_r\|_F^2$ and $V_M=\mathbb E\|\widehat C_r-K_r\|_F^2$.
Then
\begin{equation}
\mathbb E\|K_\omega-\operatorname{Cov}(H)\|_F^2
=(1-\omega)^2A_r+\omega^2V_M.
\label{eq:residualbiasvariance}
\end{equation}
When $A_r+V_M>0$, the minimizing fixed weight is $A_r/(A_r+V_M)$.
The residual fourth moment controls $V_M$, as derived in Appendix~\ref{app:residualcompletion}.
These identities describe the covariance mechanism.
The deployed weight is selected by complete calibrated loss, which also accounts for source outcomes, score scales, and calibration ranks.

The deployed rule fixes $M=8192$.
Independent complete-loss selection first chooses the degree-two ridge and scale exponent.
We retain their physical ridge $\tau_2=c_2\lambda$ and $\gamma$, then select $\omega$ from the three-point menu on the same selection pool, breaking ties toward smaller $\omega$.
The complete-operator control uses $(1-\omega)K_2+\omega\widehat C_H$ with the same states, mean, physical ridge, scale, and menu.
Both fixed-weight families have population target $K_2+\omega K_r$.

\section{Original-input numerical certification}
\label{sec:certificate}

Put $z=(1,x^\top)^\top$ and $M=\E[zz^\top H]$.
For Equation~\eqref{eq:rational},
\begin{equation}
M=M_P+\int_0^\infty G(u)\,du,\qquad
G(u)=\E[zz^\top R^2e^{-uS}].
\label{eq:matrixintegral}
\end{equation}
Every scalarization $b^\top G(u)b$ is completely monotone.
Gaussian tilting evaluates all entries at a shared node.
For $S=s_0+s^\top x+x^\top Nx$, its tilted covariance and mean are
\begin{equation}
C(u)=(I+2uN)^{-1},\qquad m(u)=-uC(u)s.
\end{equation}
Gaussian quadratic-form moment identities give $G(u)$, $\E[PR^2e^{-uS}]$, and $\E[R^4e^{-uS}]$ using $O(q^3)$ matrix work.

\begin{theorem}[Matrix and residual certificate]
\label{thm:matrix}
Positive Gauss and left-Radau rules, with original-input ball arithmetic, construct $L\preceq M\preceq U$.
Let $\widehat M=(L+U)/2$ and let
$\beta=(a_j,B_{jj}/\sqrt2,B_{jk}\ (j<k))$.
For the stored center $\widehat\beta$, define $e_{\mathrm{store}}$ to cover its distance to the exact midpoint coefficient vector and put
\begin{equation}
\norm{\beta-\widehat\beta}\leq e_\beta:=
\sqrt{(q+1)/2}\,\norm{U-L}_F/2+e_{\mathrm{store}}.
\label{eq:betabound}
\end{equation}
The mean has the associated scalar interval.
The same construction bounds $\E[H^2]$ and provides
\begin{equation}
\rho^2\leq
\overline{\E[H^2]}-\underline{\mu^2}
-\max(0,\norm{\widehat\beta}-e_\beta)^2.
\label{eq:rho}
\end{equation}
For each fixed profile and with arithmetic and storage precision chosen for the error budget, a finite geometric-panel construction attains mean and coefficient radii and a residual-energy interval width of at most $\varepsilon$ using
$O(\log(1/\varepsilon)\log\log(1/\varepsilon))$ shared nodes.
The constants retain the Schur floor and an exponential Gaussian moment.
\end{theorem}

The practical engine applies Gamma-weighted Gauss--Radau rules directly on the half-line.
The finite-panel engine implements the stated node-rate construction.
For $H^2=P^2+2PR^2/S+R^4/S^2$, the final integral uses the weight $u\,du$.
Its tail is bounded by
\begin{equation}
\int_U^\infty u\E[R^4e^{-uS}]\,du
\leq (U/\delta+1/\delta^2)\E[R^4e^{-US}].
\end{equation}
Gauss--Radau error signs, Gaussian moments, and ball arithmetic retain their classical attribution \citep{gautschi2004,johansson2018}.

\paragraph{Propagation to the numerical predictor.}
We form an interval lift using the certified $L_i$ and compare it with the actual stored global point feature $\gamma_i^p$.
Isometry gives $e_i=e_i^{\mathrm{loc}}+e_i^{\mathrm{lift}}$.
Use the common response unit $T>0$ and let $K_c,k_c$ be the exact Gram and cross kernel for $H/T$, divided by the same frozen normalizer $c>0$.
Let $d_K,d_k$ bound their errors relative to $\widehat K_c,\widehat k_c$, and let $a_i$ bound point-mean errors in units of $H/T$.
For the actual saved coefficient vector $\widetilde\alpha$ and $\lambda>0$, put
$r=Y/T-\mu^p-(\widehat K_c+\lambda I)\widetilde\alpha$.
For any certified $C_*\geq\norm{(K_c+\lambda I)^{-1}k_c}$,
\begin{equation}
\begin{split}
|p_*^{\mathrm{exact}}-p_*^{\mathrm{saved}}|/T
\leq{}&a_*+d_k\norm{\widetilde\alpha}+d_{\mathrm{dot}}\\
&+C_*(\norm r+\norm{a_{\mathrm{src}}}+d_K\norm{\widetilde\alpha}).
\end{split}
\label{eq:deploybound}
\end{equation}
The gain bound combines the Gram singular-value bound with a certified dual-residual bound.
All quantities use a common response unit and the actual frozen normalizer.
Appendix~\ref{app:deployment} gives the complete formulas and the anchored affine-extension risk.

For Equation~\eqref{eq:residualkernel}, the exact finite-state target has feature blocks $(g_i,\sqrt\omega u_i)$.
We enclose the original Schur coefficients once per task, then bound all sample evaluations with outward arithmetic.
Let $\zeta_i\in\mathbb R^{d_i}$ collect the task's active private and shared primitive Gaussian coordinates.
For the normalized first- and second-order Hermite features, excluding the constant, the identity $\|\Psi_2(\zeta_i)\|_2^2=(\|\zeta_i\|_2^4+d_i)/2$ converts coefficient radii to pointwise polynomial errors.
Centering removes the taskwise constant mean error.
Separate low-order and residual Gram perturbations provide $d_K,d_k$ in Equation~\eqref{eq:deploybound}.
This certifies the saved predictor at its actual states and selected hyperparameters.
Appendix~\ref{app:residualnumerics} gives the full error chain.

\section{Calibrated prediction and exact task risk}
\label{sec:calibration}

After source fitting and independent candidate selection, freeze $p(X)$ and a positive score scale $s(X)$.
For calibration tasks, compute $S_i=(Y_i-p_i)/s_i$.
With $m$ calibration scores and $k=\lceil .95(m+1)\rceil$, return
\begin{equation}
U_*(\mathcal C)=\max\{0,p_*+s_*S_{\mathcal C,(k)}\}.
\label{eq:upper}
\end{equation}

\begin{theorem}[Task-level coverage]
\label{thm:coverage}
For $1\leq k\leq m$, conditional on a fixed reference and frozen learning procedure, exchangeable calibration and test task scores satisfy
$\Pr\{Y_*\leq U_*(\mathcal C)\}\geq k/(m+1)$.
For a fixed finite score pool of size $N>m$, the same lower bound holds when the calibration subset is uniformly sampled and the test task is uniformly sampled from its complement.
With $m=256,k=245$, the bound is $245/257$.
\end{theorem}

We evaluate the complete finite-pool risk
\begin{equation}
D=\E_{\mathcal C}\left[
\frac1{N-m}\sum_{i\notin\mathcal C}\loss(Y_i,U_i(\mathcal C))\right],
\label{eq:risk}
\end{equation}
where $\loss(y,u)=\max\{.95(y-u),.05(u-y)\}$.
The split variance $V$ is the variance of the bracketed batch loss over calibration subsets.

\begin{theorem}[Complete finite-pool calculation]
\label{thm:finiterisk}
For fixed nonnegative labels $Y_i$, predictions $p_i$, positive scales $s_i$, and $1\leq k\leq m<N$, the complete risk $D$ and split variance $V$ can be computed in $O(N\log N)$ real-arithmetic time and $O(N)$ storage.
The calculation uses hypergeometric pivot weights and sorted prefixes at the score and clipping breakpoints.
\end{theorem}

Appendix~\ref{app:finiterisk} gives the conditional moment formulas and the two-sort construction.
For simultaneous prediction radii $\epsilon_i$, order $a_i=\epsilon_i/s_i$ increasingly and carry the scales with that order.
Let $v_j=\binom{j-1}{m-1}/\binom Nm$, $S_j^-=\sum_{i<j}s_{(i)}$, and $S_j^+=\sum_{i>j}s_{(i)}$.
The complete calibration perturbation is
\begin{equation}
\begin{split}
|D(\widetilde p)-D(p)|\leq .95\bigl(\bar\epsilon+\Phi_m\bigr),\\
\Phi_m=\sum_{j=m}^N\frac{v_ja_{(j)}}{N-m}
\left(S_j^++\frac{j-m}{j-1}S_j^-\right).
\end{split}
\label{eq:sharprisk}
\end{equation}
The second term in parentheses is zero for $j=m=1$.
Conditioning on the largest calibration rank proves the formula, and sorted prefixes evaluate it in $O(N\log N)$ time.
Combining Equation~\eqref{eq:sharprisk} with an Arb enclosure of the saved risk gives an interval for the exact finite-state predictor's risk.
Appendix~\ref{app:residualcompletion} supplies the proof and exact-enumeration checks.
For reference-fitted task scales $s_i$, the independent selection pool jointly chooses ridge and a scale exponent $\gamma\in\{0,1/2,1\}$ using this complete $D$.
The score scale is $(s_i/s_{\mathrm{ref}})^\gamma$, where $s_{\mathrm{ref}}$ is the source geometric mean.
The two endpoints of the menu recover the unit and common scales.
Appendix~\ref{app:scalelearning} gives a deterministic grid-approximation bound.
An independent evaluation pool measures the frozen rule.
The complete calculation averages every calibration subset and its complementary recipients.
Its U-statistic form also gives exact multiscale subsampling identities and a delete-one sensitivity statistic, derived in Appendix~\ref{app:subsampling}.

\section{Experiments}
\label{sec:experiments}

\begin{algorithm}[t]
\caption{Functional prediction and task calibration}
\label{alg:method}
\begin{algorithmic}[1]
\REQUIRE Known task operators, partial row context, source outcomes, selection outcomes, calibration outcomes
\STATE Freeze row centers, scales, row features, and the shared row law.
\FOR{each source, selection, calibration, and query task}
\STATE Merge repeated row identities and whiten its local coordinates.
\STATE Compute its mean and complete degree-two projection from shared tilted Gaussian moments.
\STATE Lift the coefficients into the common sparse feature space.
\ENDFOR
\STATE Form the degree-two kernel and select its ridge and score-scale exponent by complete calibrated loss.
\STATE Evaluate 8192 shared states and form centered residual covariance features.
\STATE Retain the selected physical ridge and score scale, and select the residual weight from $\{0,1/2,1\}$.
\STATE Freeze the predictor and score scale, then compute the calibration rank endpoint.
\STATE For a numerical certificate, propagate original-input and sample enclosures through the saved solve, point product, and complete-risk calculation.
\STATE \textbf{return} Upper endpoint and the requested numerical certificate
\end{algorithmic}
\end{algorithm}

\paragraph{Reference tables and task protocol.}
The fixed panel contains Gaussian, heteroscedastic, heavy-tailed, and nonlinear simulations, together with Concrete, Airfoil, Power Plant, white Wine, red Wine, and Superconductivity regression tables.
For each reference, training-only standardization and ridge fitting define a candidate.
The feature with the largest fitted coefficient magnitude becomes the shared hidden column.
Tasks draw $q\in\{8,32\}$ rows with replacement and retain the 64 observed anchors and all remaining design columns.
Each anchor and each sampled occurrence has objective weight $1/(64+q)$.
Repeated draws share their hidden value and contribute their summed occurrence weights.

The source fitting budget is 640 scalar regret outcomes.
An independent 512-task pool selects from $\lambda\in\{10^{-8},\ldots,10^1\}$ and $\gamma\in\{0,1/2,1\}$.
Each evaluation split uses 256 calibration outcomes.
Deployment thus uses 1,408 observed outcomes across these three roles.
\EvaluationProtocolSentence{}
Each simulated reference redraws its table, and each measured-table reference permutes and partitions the same public records before drawing new tasks.
All ten benchmark populations enter every aggregate.
Per-dataset results first average the original-unit $D$ across both task sizes and all listed reference seeds.
Dataset ratios then receive equal weight.

\paragraph{Public comparisons.}
The comparison includes linear quantile regression, ridge regression, random forests, Extra Trees, quantile regression forests, histogram quantile boosting, LightGBM quantile regression, and the official density-weighted CPCP network with vanilla and Clip+Mix training.
We also evaluate bag kernels, an input RBF kernel, and a random-forest proximity kernel.
Direct ensemble linear-Bayes controls evaluate the complete operators on shared hidden-state samples, with three sampling budgets and two mean offsets.
Every family receives the same source-outcome budget, and the generic bag controls also receive the same row identities and row kernel.
Every family receives the same three-point score-scale menu and independent complete-loss selection budget.
The CPCP adapter retains its official fixed architecture and finite-difference density weighting.
The appendix gives its single-sided response and budget adaptation \citep{koenker1978,breiman2001,geurts2006,meinshausen2006,ke2017,chen2026}.

\paragraph{Evaluation on new reference contexts.}
The final evaluation uses references 5501--5520, both task sizes, and 2048 evaluation tasks per condition.
The mean dataset risk ratio to the degree-two kernel is 0.99768, with a 95\% paired reference-bootstrap interval of [0.99632,0.99881].
Point estimates improve on seven datasets, with interval upper bounds below one on five.
The incremental reductions are 1.75\% on Heavy-tailed and 0.47\% on Airfoil. The remaining dataset ratios range from 0.99868 to 1.00078.
The corresponding mean ratio to matched 8192-state complete-MC shrinkage is 0.99610, with interval [0.99520,0.99699].
The ratio to the retrospective public minimum is 0.99109, with interval [0.98978,0.99275], and is below one on all ten datasets.
The fixed-RF ratio is 0.69602.
\begin{table}[t]
\caption{Residual-rule complete-loss ratios on twenty new reference contexts per dataset. Each dataset averages original-unit loss across both task sizes and all contexts. Public min is the retrospective minimum over 174 public configurations.}
\label{tab:confirmation}
\centering
\small
\setlength{\tabcolsep}{4pt}
\begin{tabular}{lrrr}
\toprule
Dataset & $K_2$ & Matched MC & Public min\\
\midrule
Gaussian & 0.999877 & 0.9959 & 0.9927\\
Heteroscedastic & 0.999773 & 0.9973 & 0.9925\\
Heavy-tailed & 0.982489 & 0.9884 & 0.9908\\
Nonlinear & 0.999966 & 0.9969 & 0.9941\\
Concrete & 1.000782 & 0.9984 & 0.9846\\
Airfoil & 0.995314 & 0.9957 & 0.9882\\
Power Plant & 0.999712 & 0.9972 & 0.9943\\
Wine white & 1.000170 & 0.9987 & 0.9920\\
Wine red & 0.998685 & 0.9961 & 0.9935\\
Superconductivity & 1.000005 & 0.9965 & 0.9882\\
\bottomrule
\end{tabular}
\end{table}

\begin{figure}[t]
\centering
\includegraphics[width=\columnwidth]{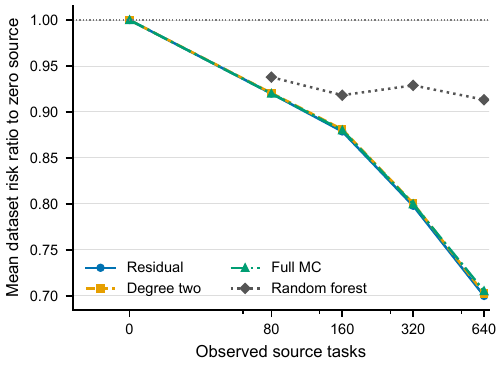}
\caption{Learning from scalar source observations on the fixed ten-dataset development panel. Complete risk is normalized by the zero-source analytic-mean risk within each dataset and then averaged. Full MC uses the matched 8192-state shrinkage rule. The selection budget is 512 outcomes at every source budget.}
\label{fig:sourcelearning}
\end{figure}

\paragraph{Learning from source observations.}
We fix the working law, task context, shared states, selection pool, and evaluation pool on all ten datasets at development reference 4301.
Only the nested source budget changes over $n\in\{0,80,160,320,640\}$.
At zero source observations the predictor is the analytic mean, with the same three-point scale selection and calibration procedure.
The positive-budget kernel rules select their parameters independently at each budget.
Figure~\ref{fig:sourcelearning} shows the equal-dataset average of $D(n)/D(0)$.
For the residual rule it is $.9200,.8788,.7980,.6999$ at the four positive budgets.
The matched complete-MC shrinkage control uses the same 8192 states, degree-two target, physical ridge, scale, and weight menu.
The residual-to-control ratios are $1.00013,.99845,.99757,.99255$.
The appendix gives every dataset and both task sizes, together with the 32768-state full-MC comparison.

\IfFileExists{covariance_control_main_results.tex}{\paragraph{Same-sample covariance control.}
A direct control-variate adaptation uses $K_2+\omega(\widehat C_H-\widehat C_{F_2})$ and a positive-source eigenspace with a consistent query extension.
It receives the same physical ridge, scale, states, and weight-selection menu as the residual rule.
On all 80 development conditions with four streams, the mean dataset residual-to-control loss ratio is 0.99767, with lower point estimates on 8 of ten datasets.
Appendix~\ref{app:covariancecontrol} gives every dataset, sampling-variation comparison, and spectrum diagnostic.
}{}
\IfFileExists{matched_family_cost_main_results.tex}{\paragraph{Matched production cost.}
A fresh-process study recomputes all three matched covariance families on eight fixed development conditions, including their common degree-two construction and base selection.
The residual rule's geometric mean production-time ratio is 1.197 to complete-MC shrinkage and 0.942 to the positive-source control-variate adaptation.
The measured components include construction, sampling, fitting, prediction, selection, and complete-risk evaluation. Appendix~\ref{app:matchedcost} gives each condition and peak process memory.
}{}

\paragraph{Certified precision and computation cost.}
The coefficient computation is compared on 40 profiles at tolerances $10^{-5},10^{-7},10^{-9}$ against reliable Arb/Petras integration with shared tilted moments and real-midpoint preconditioning \citep{johansson2018}.
All three engines meet the same mean, coefficient-radius, and residual-energy accuracy requirements.
The Gamma engine has lower cold time on every profile at every tolerance.
At $10^{-7}$, its geometric mean cold-time ratio to the strengthened Petras adapter is .0646.
Appendix~\ref{app:numericalexperiments} reports profile-level accuracy and cost.
For the finite-state residual predictor, three complete development pools contain 3456 source and query functions and nine fixed-weight branches.
Their complete-risk error radii range from $2.762\times10^{-5}$\% to $3.072\times10^{-3}$\% of the computed risk.
Appendix~\ref{app:residualnumerics} specifies the conditioning quantities, coefficient precision, sample-value bounds, and computation cost.

\section{Related work}

Distribution regression embeds bags before learning a response, and aggregate-output Gaussian processes share latent instance functions across overlapping bags.
Learning from aggregate observations also studies likelihoods and identifiability for grouped supervision.
These provide direct context for the shared-instance observation model \citep{szabo2016,law2018,zhang2020}.

Polynomial-chaos Kalman filtering updates chaos coefficients through covariance operators, and linear conditional-mean prediction has the covariance form used in Equation~\eqref{eq:prediction}.
PCEGP uses polynomial expansions for input-dependent Gaussian-process hyperparameters.
Our computational object is the projection of a known rational task operator, its original-input coefficient balls, and their propagation through a shared sparse task kernel \citep{ernst2015,vondrejc2018,polke2024}.

Polynomial-chaos control variates combine analytic surrogate statistics and Monte Carlo corrections \citep{fox2021,yang2023}.
CVPC analyzes the allocation between surrogate order and high-fidelity sampling, including separate control weights for mean and variance estimation \citep{yang2023}.
Multi-fidelity covariance estimation in log-Euclidean geometry preserves positive definiteness and provides fixed-budget sample allocation \citep{maurais2023}.
Our common task features support the incremental query blocks and the original-input error chain required by the saved calibrated predictor.

Conformalized quantile regression and density-weighted conformal learning motivate task-level upper bounds and strong learned score baselines.
Conformal Risk Training differentiates through conformal OCE risk control to train models with downstream risk feedback \citep{yeh2025}.
Our scale selection uses exact averaging over calibration roles on an independent finite task pool.
Here the operator kernel supplies the point predictor, and independent task-rank calibration supplies the finite-sample guarantee.
Positive Gaussian quadrature and arbitrary-precision ball arithmetic supply the numerical certification primitives \citep{romano2019,chen2026,gautschi2004,johansson2018}.

Moments of Gaussian quadratic-form ratios have integral, series, and polynomial-recursion formulations, with numerical implementations for ratio and product moments \citep{bao2013,watanabe2023}.
The computation encloses all signed projection coefficients in one positive matrix integral and propagates its error through sparse assembly to the saved predictor.

\section{Conclusion}

Known nonlinear audit operators and shared row identities support functional prediction from scalar task observations.
Local projections and shared residual features yield a common kernel for new audit functions.
Original-input and sample enclosures propagate through the saved predictor to complete calibrated loss.
The experiments quantify the incremental residual-covariance gain, learning from source observations, and the precision--cost relation of certified computation.

\clearpage
\typeout{CONTENT_PAGES:\number\numexpr\value{page}-1\relax}
\section*{AI Disclosure Statement}
Generative AI assisted ideation, mathematical reasoning, experimental design, code, auditing, literature review, data analysis, and drafting.
Code, experiment records, and numerical verification results accompany the paper.
The authors retain responsibility for all content.

\section*{Impact Statement}
The method supports repeated scalar audits of partially observed training tables.
Experiments use public regression records with artificial feature withholding and controlled simulations.
Operational use requires the task population and calibration roles stated here.

\clearpage
\onecolumn
\appendix
\section{Task operators and the working distribution}
\label{app:operators}

\subsection{Positive Schur representation}

The algebraic reduction follows the ridge-regret representation of \citet{li2026}.
For a completed design $X=[A,z]$, define
\[
 B=A^\top WA+\lambda_0I,\quad a=A^\top Wy,\quad
 c=A^\top Wz,\quad b=z^\top Wy,
\]
\[
 S=z^\top Wz+\lambda_0-c^\top B^{-1}c,\qquad r_0=b-c^\top B^{-1}a.
\]
Write the fixed candidate as $(\beta_A,\beta_z)$.
Completing the square first in the observed coefficients and then in the hidden coefficient gives
\[
 H=P+R^2/S,
\quad
 P=\left\|B^{1/2}(\beta_A-B^{-1}a+B^{-1}c\beta_z)\right\|^2,
\quad R=\beta_zS-r_0.
\]
When $z$ is affine in the hidden Gaussian coordinates, $P,R,S$ have degree at most two and $P\geq0$.
The Schur complement of the positive ridge matrix satisfies $S\geq\lambda_0$.
Indeed, it is the minimum over $u$ of $(Au+z)^\top W(Au+z)+\lambda_0\|u\|^2+\lambda_0$.
This supplies an original-input positive floor for the integrals.
All observed columns, the intercept, the anchor rows, and the actual task weights enter this calculation.

\subsection{Shared row coordinates}

Let $P_i$ now denote the row-incidence map after merging repeated cohort identities.
In primitive coordinates $Z=(\epsilon,g)$, the standardized row values selected by a task have map
\[
 T_i=[\sqrt{1-\eta}P_i,\sqrt\eta\Phi_i],\quad
 L_iL_i^\top=T_iT_i^\top,\quad A_i=L_i^{-1}T_i.
\]
Here $\Phi_i$ contains the normalized features of the distinct selected rows.
Consequently $A_iA_i^\top=I$.
The original row scales and centers are applied before the operator coefficients are formed.
Repeated appearances of a row use the same center, scale, and primitive coordinate.

Row features concatenate the observed design row, observed response, predicted hidden-feature center, and log scale.
Their centering and coordinate scales use the distinct rows appearing in the 640 source tasks.
Numerically constant coordinates are removed, an intercept is added, and each feature vector is normalized to unit length.
The retained dimension is denoted by $d$.
Rows that occur only in later task inputs are transformed with these frozen constants.
The common score scale is $s_i=\sqrt{t_i^{(C)}t_i^{(R)}}$, where the two auxiliary quadratic-Taylor standard deviations are defined in Appendix~\ref{app:workingquantities}.
All these quantities are computed from the allowed partial-table context.

\section{Proof of the local-to-global kernel}
\label{app:kernelproof}

For standard Gaussian $Z$, the first- and centered second-order chaoses are orthogonal.
For vectors $u,v$ and symmetric matrices $C,D$,
\[
 \E[(u^\top Z)(v^\top Z)]=u^\top v,\qquad
 \E[(Z^\top CZ-\tr C)(Z^\top DZ-\tr D)]=2\tr(CD).
\]
Substitute $x_i=A_iZ$ in Equation~\eqref{eq:projection}.
The lifted coefficients are $A_i^\top a_i$ and $A_i^\top B_iA_i$.
The identities above yield Equation~\eqref{eq:kernel}.
Taking $i=j$ and using $A_iA_i^\top=I$ proves the isometry.

To identify the global residual, consider any polynomial $Q(Z)$ of total degree at most two.
The conditional expectation $\E[Q(Z)\mid A_iZ]$ is a polynomial of degree at most two in $A_iZ$.
Since the local residual is orthogonal to every such polynomial,
\[
 \E[r_iQ(Z)]=\E[r_i\E[Q(Z)\mid A_iZ]]=0.
\]
Thus the lifted local projection is the global degree-two orthogonal projection of the task function.
Every cross term between a projected function and a residual vanishes.
The matrix $K_r=(\E[r_ir_j])_{ij}$ is a Gram matrix in $L^2$ and is positive semidefinite.
Cauchy--Schwarz gives $|(K_r)_{ij}|\leq\rho_i\rho_j$.

\subsection{Sparse common features}

In the original selected-row coordinates, write a task's centered polynomial as
$u_i^\top\xi_i+(\xi_i^\top V_i\xi_i-\tr(V_iR_{ii}))/2$.
The global coefficient vector contains the following entries with common cohort indices.
\[
\begin{array}{ll}
\text{private linear} & \sqrt{1-\eta}\,P_i^\top u_i,\\
\text{private quadratic} & (1-\eta)P_i^\top V_iP_i,\\
\text{shared linear} & \sqrt\eta\,\Phi_i^\top u_i,\\
\text{shared quadratic} & \eta\Phi_i^\top V_i\Phi_i,\\
\text{private--shared} & \sqrt{\eta(1-\eta)}P_i^\top V_i\Phi_i.
\end{array}
\]
Quadratic diagonal entries are divided by $\sqrt2$ and off-diagonal entries are stored once.
These weights implement the norm $\|a\|^2+\|B\|_F^2/2$.
The private blocks and private--shared block are sparse.
For $n$ sources and $Q$ queries, local coefficient storage is
$O((n+Q)(q^2+qd+d^2))$ for a common task size $q$.
The dense source kernel uses $O(n^2)$ storage, and the present exact ridge solver uses $O(n^3)$ time.
Kernel assembly uses sparse common-coordinate products and dense shared-coordinate products.

\section{Original-input matrix and residual certificates}
\label{app:matrixproof}

\subsection{Positive matrix integral}

For $S\geq\delta>0$, Tonelli's theorem applied to the nonnegative scalarization gives
\[
 M=M_P+\int_0^\infty G(u)\,du,\qquad
 G(u)=\E[zz^\top R^2e^{-uS}],\quad z=(1,x^\top)^\top.
\]
Differentiation under the Gaussian integral gives, for every vector $v$ and integer $j\geq0$,
\[
 (-1)^jv^\top G^{(j)}(u)v=\E[(v^\top z)^2R^2S^je^{-uS}]\geq0.
\]
Hence every scalarization is completely monotone.
For a finite interval, the Gaussian quadrature remainder has a nonnegative even derivative.
The left-endpoint Radau remainder has a nonpositive odd derivative.
Their scalar lower and upper inequalities for every $v$ prove the Loewner inequalities.
The rules and their remainder signs follow classical positive quadrature theory \citep{gautschi2004}.

The practical half-line implementation uses $t=\delta u$ and
\[
 F(t)=\E[zz^\top R^2e^{-t(S/\delta-1)}].
\]
Let $(t_j^{(a)},w_j^{(a)})$ be the $n$-node Gauss rule for the normalized Gamma distribution with shape $a$.
Then
\[
 M_P+\delta^{-1}\sum_jw_j^{(1)}F(t_j^{(1)})\preceq M
\preceq
 M_P+\delta^{-1}\left\{\frac{F(0)}{n+1}
 +\sum_j\frac{w_j^{(2)}}{t_j^{(2)}}F(t_j^{(2)})\right\}.
\]
The right rule is the positive left-Radau rule for the exponential measure.
Its zero-node weight is $1/(n+1)$ and its other nodes are the Gamma-shape-two Gauss nodes.
The Gamma moment problem is determinate.
Positive quadrature measures with increasing matched moments converge weakly to that measure.
The bounded continuous function $F$ therefore has matching limits under both rules.
This establishes convergence of the enclosure for a fixed profile.

\subsection{Shared Gaussian tilt}

Write $R=r_0+r^\top x+x^\top Ex$ and $S=s_0+s^\top x+x^\top Nx$.
At a real node $u\geq0$, define
\[
 C=(I+2uN)^{-1},\quad m=-uCs,\quad
 Z_u=\det(I+2uN)^{-1/2}\exp(-us_0+u^2s^\top Cs/2).
\]
Let $\bar r=r_0+r^\top m+m^\top Em+\tr(EC)$ and $\ell=r+2Em$.
Under the tilted Gaussian $\Normal(m,C)$, put
\[
 s_2=\bar r^2+\ell^\top C\ell+2\tr((EC)^2),\qquad
 h=2C(\bar r\ell+2EC\ell),
\]
\[
 Q=C(2\ell\ell^\top+8ECE+4\bar rE)C.
\]
Gaussian integration by parts gives
\[
 G(u)=Z_u\begin{pmatrix}
 s_2 & (ms_2+h)^\top\\
 ms_2+h & (C+mm^\top)s_2+mh^\top+hm^\top+Q
 \end{pmatrix}.
\]
This evaluates every coefficient moment from the same tilt.
For $P=p_0+p^\top x+x^\top Q_Px$, its polynomial contribution is
\[
 M_P=\begin{pmatrix}\bar p&p^\top\\p&\bar pI+2Q_P\end{pmatrix},\qquad
 \bar p=p_0+\tr Q_P.
\]
All these operations use arbitrary-precision balls built from the exact real interpretation of the original binary64 inputs.

\subsection{Extracting a Euclidean coefficient ball}

Let $E=M-(L+U)/2$ and $F=(U-L)/2$.
Both $F-E$ and $F+E$ are positive semidefinite, so
\[
 0\leq\tr((F-E)(F+E))=\|F\|_F^2-\|E\|_F^2.
\]
The linear map from a symmetric moment matrix to
$\beta=(a_j,B_{jj}/\sqrt2,B_{jk}\ (j<k))$ has Frobenius-to-Euclidean norm $\sqrt{(q+1)/2}$.
On the joint coordinates $(E_{00},\tr(E_{1:q,1:q})/\sqrt q)$, the map has scalar component
$(\tr(E_{1:q,1:q})/\sqrt q-\sqrt qE_{00})/\sqrt2$ and squared operator gain $(q+1)/2$.
The remaining orthogonal directions have squared gain at most $1/2$.
Combining these facts proves Equation~\eqref{eq:betabound} before storage.
The implementation adds the norm of the outward-rounded storage-error vector.
Equivalently, $\beta^0=\mathcal T((L+U)/2)$ and
$\|\widehat\beta-\beta^0\|\leq e_{\mathrm{store}}$ define the stored center and radius used in the residual calculation.

An entrywise symmetric ball matrix is converted to a Loewner ball by taking its exact midpoint and adding or subtracting an upper bound on the entry-radius matrix's maximum row sum times the identity.
This accounts for original-input elimination, whitening, nodes, weights, integration, and arithmetic.

\subsection{Second moment and higher-order energy}

The positive decomposition
\[
 H^2=P^2+2PR^2/S+R^4/S^2
\]
uses exact polynomial Gaussian moments for the first term.
The second term uses the same shape-one Gauss and left-Radau rules.
For the final term, the shape-two measure has a left-Radau upper rule with zero weight
$2/((n+1)(n+2))$ and positive-node weights $2w_j^{(3)}/t_j^{(3)}$.

The centered cumulants of $R$ under a tilt are
\[
 \kappa_2=\ell^\top C\ell+2\tr((EC)^2),\quad
 \kappa_3=6\ell^\top CEC\ell+8\tr((EC)^3),
\]
\[
 \kappa_4=48\ell^\top CECEC\ell+48\tr((EC)^4).
\]
Thus $\E[R^4]=\bar r^4+6\bar r^2\kappa_2+3\kappa_2^2+4\bar r\kappa_3+\kappa_4$.
The tilted polynomial mean is
$\bar p_u=p_0+p^\top m+m^\top Q_Pm+\tr(Q_PC)$.
The matrices already used for $G$ also give the original integral's scalar integrand
$\E[PR^2e^{-uS}]=Z_u\{\bar p_us_2+(p+2Q_Pm)^\top h+\tr(Q_PQ)\}$.
Projection orthogonality gives
$\rho^2=\E[H^2]-\mu^2-\|\beta\|^2$.
Using interval endpoints and $\|\beta\|\geq\max(0,\|\widehat\beta\|-e_\beta)$ proves Equation~\eqref{eq:rho}.

\subsection{Finite-panel node complexity}
\label{app:complexity}

Choose $a>0$ with $2a\|N\|_2<1$.
All quadratic matrices use symmetric representations, and $S\geq\delta$ implies $N\succeq0$.
All polynomial Gaussian exponential moments needed for $G$, $PR^2$, and $R^4$ are then finite at exponent $aS$.
The tilted integrands extend analytically to $\operatorname{Re}u>-a$.
Their matrix or scalar magnitudes are bounded there on $\operatorname{Re}u\geq-a/2$ by corresponding exponential moments.

Choose $U=O(\log(1/\varepsilon))$ using the positive tails
\[
 \int_U^\infty G(u)\,du\preceq G(U)/\delta,
\qquad
 \int_U^\infty\E[PR^2e^{-uS}]\,du\leq\E[PR^2e^{-US}]/\delta,
\]
\[
 \int_U^\infty u\E[R^4e^{-uS}]\,du
 \leq(U/\delta+1/\delta^2)\E[R^4e^{-US}].
\]
The exponential factor $e^{-\delta U}$ dominates their polynomial factors for a fixed profile.
Partition $[0,U]$ into $[b,\min(2b+a,U)]$.
There are $O(\log(1+U/a))$ panels.
The Bernstein ellipse of parameter two around each panel has real part at least $-a/8$.
Chebyshev approximation of degree $2n-1$ on each panel consequently has error bounded by a profile constant times $4^{-n}$.
Positive quadrature weights preserve the uniform approximation bound.
For the $uR^4$ term, split $u=b+(u-b)$ and use the positive uniform and linear-weight rules.
Combining the positive panel rules and tails bounds the exact-arithmetic enclosure gap by
\[
 C_1(U+U^2)4^{-n}+C_2(1+U)e^{-\delta U},
\]
with profile-dependent finite constants $C_1,C_2$.
Taking $n=O(\log(1/\varepsilon))$ makes this expression smaller than the specified matrix and scalar budgets.
The coefficient map is linear, and the mean square and coefficient-norm square are locally Lipschitz at a fixed profile.
Rescaling those budgets by their finite profile constants bounds the mean radius, coefficient radius, and residual-energy interval width by $\varepsilon$.
In particular, if $w_2,w_\mu$ are the second-moment and mean interval widths, and $M_\mu$ bounds the mean interval's absolute endpoints, then
\[
 \operatorname{wid}(I_{\rho^2})\leq
 w_2+2M_\mu w_\mu+4(\|\widehat\beta\|+e_\beta)e_\beta.
\]
The shared node count is therefore
$O(\log(1/\varepsilon)\log\log(1/\varepsilon))$.
The constants depend on the fixed profile, $\delta$, $a$, dimension, and exponential moments.
Arithmetic precision is increased until the outward-rounding budgets are met.
The asymptotic construction also allows the coefficient-center storage precision to increase with the requested tolerance.
The deployed binary64 implementation includes its computed storage radius in the reported contracts.
This is a node-count result for the finite-panel construction.
The half-line implementation uses its computed enclosure as a stopping rule.

\section{Certificate for the saved prediction}
\label{app:deployment}

All following quantities are expressed in one common response unit $T$.
The experiment computes means and features for $H/T$.
Let $c>0$ be the actual stored normalizer for these features, let $\lambda>0$, and let
$K=\Gamma\Gamma^\top/c$ and $k=\Gamma\gamma_*/c$ denote the exact working-kernel matrices.
Their saved floating-point versions are $\widehat K,\widehat k$.
Let $\gamma_i^p$ be the actual point feature and $e_i$ a certified radius around it.
Set
\[
 s^2=\sum_i\|\gamma_i^p\|^2/c,\qquad e^2=\sum_i e_i^2/c.
\]
If $A_K,A_k$ cover assembly arithmetic for those point features, then
\[
 d_K=2se+e^2+A_K,
\]
\[
 d_k=\frac{\sqrt{\sum_i\|\gamma_i^p\|^2}\,e_*+
 \sqrt{\sum_i e_i^2}(\|\gamma_*^p\|+e_*)}{c}+A_k
\]
cover the operator and cross-vector errors.
The local radius and the certified lift discrepancy are added using the isometry.
The point-mean error is the maximum distance to either certified mean endpoint.

For verified normal-range binary64 dot products, a bound with
$\gamma_{L+32}=(L+32)2^{-53}/(1-(L+32)2^{-53})$ covers the block products, block sums, symmetrization, and normalization.
Here $L$ bounds the number of nonzero terms in a row.
Cauchy--Schwarz gives $A_K\leq\gamma_{L+32}s^2$ and
$A_k\leq\gamma_{L+32}s\|\gamma_*^p\|/\sqrt c$.

Let $y=Y/T$ and let $\widetilde\alpha$ be the actual saved coefficient vector.
Define
\[
 r=y-\mu^p-(\widehat K+\lambda I)\widetilde\alpha.
\]
For $\alpha=(K+\lambda I)^{-1}(y-\mu)$,
\[
 (K+\lambda I)(\alpha-\widetilde\alpha)
 =r+(\mu^p-\mu)+(\widehat K-K)\widetilde\alpha.
\]
Taking its inner product with $(K+\lambda I)^{-1}k$ and bounding the direct cross-kernel term proves
\[
 |p_*^{\mathrm{exact}}-p_*^{\mathrm{saved}}|/T
 \leq a_*+d_k\|\widetilde\alpha\|+d_{\mathrm{dot}}
 +C_*(\|r\|+\|a_{\mathrm{src}}\|+d_K\|\widetilde\alpha\|).
\]
The term
$d_{\mathrm{dot}}=|p_*^{\mathrm{saved}}/T-(\mu_*^p+\widehat k^\top\widetilde\alpha)|$
compares the exact dot product of the saved floating-point operands with the saved scalar prediction.

For the gain, a singular-value decomposition of $\Gamma/\sqrt c$ gives
\[
 \|(K+\lambda I)^{-1}k\|
 \leq (\|\gamma_*^p\|+e_*)/(2\sqrt{c\lambda}).
\]
For any approximate source weight $w_0$, define
$r_0=\widehat k-(\widehat K+\lambda I)w_0$.
The resolvent bound $\|(K+\lambda I)^{-1}\|\leq1/\lambda$ also gives
\[
 \|(K+\lambda I)^{-1}k\|
 \leq\|w_0\|+(\|r_0\|+d_k+d_K\|w_0\|)/\lambda.
\]
The certificate uses the smaller of these two certified upper endpoints.
Thus $C_*=\min(C_{\mathrm{SVD}},C_{\mathrm{dual}})$.
All residuals and norms used here are recomputed with directed rounding.
The original-unit ridge is $T^2c\lambda$.

\subsection{Anchored affine-extension risk}

Freeze $w_0$, the observed source labels, the selected ridge, row features, and normalizer.
In this subsection $\gamma_i^p,e_i,a_i,\rho_i$ all refer to $H_i/T$.
Choose $b$ so that $b+w_0^\top(Y/T)=p_*^{\mathrm{saved}}/T$ at these observed labels.
Define the affine extension $\widetilde H_*(Z)/T=b+\sum_iw_{0i}H_i(Z)/T$.
If $d_b$ bounds the discrepancy between $b$ and
$\mu_*^p-\sum_iw_{0i}\mu_i^p$, orthogonality and the triangle inequality give
\[
 \E[(H_*(Z)-\widetilde H_*(Z))^2]/T^2
 \leq
 (a_*+\sum_i|w_{0i}|a_i+d_b)^2
 +(\|\gamma_*^p-\sum_iw_{0i}\gamma_i^p\|+e_*+\sum_i|w_{0i}|e_i)^2
 +(\rho_*+\sum_i|w_{0i}|\rho_i)^2.
\]
The integration variable is a fresh unconditional $Z\sim\Normal(0,I)$ after these quantities are frozen.
The saved-prediction numerical error above and this affine-function risk have their respective stated objects.

\section{Coverage and complete finite-pool loss}
\label{app:finiterisk}

\subsection{Rank coverage}

Condition on the reference and the frozen learning outputs.
For exchangeable calibration and test scores, the test rank among $m+1$ scores is uniform after random tie breaking.
At least $k$ ranks satisfy $S_*\leq S_{\mathcal C,(k)}$.
Using weak inequalities for the original tied scores preserves the lower bound.
The nonnegative clipping of the upper endpoint preserves coverage for $Y_*\geq0$.
This is the one-sided rank argument underlying split conformal prediction \citep{romano2019}.

For a fixed finite pool, first draw an unordered set of $m+1$ indices uniformly.
Then designate one of these indices uniformly as the test index.
The remaining $m$ indices form a uniform calibration subset and the test is uniform over its complement.
The preceding rank count proves the finite-pool assertion for every fixed score vector.

\subsection{Exact risk and variance}

Sort the $N$ scores stably and index their ranks by $j$.
The probability that rank $j$ is the $k$th calibration score is
\[
 w_j=\frac{\binom{j-1}{k-1}\binom{N-j}{m-k}}{\binom Nm},
 \qquad k\leq j\leq N-m+k.
\]
Conditional on that pivot, the lower complement is a uniform set of
$h_L=j-k$ recipients from $M_L=j-1$ lower ranks.
The upper complement is an independent uniform set of
$h_U=N-j-m+k$ recipients from $M_U=N-j$ upper ranks.
Put $d=N-m$ and $\ell_{ij}=\loss(Y_i,\max(0,p_i+s_iS_j))$.
For each side $B\in\{L,U\}$, define $A_B=\sum_{i\in B}\ell_{ij}$ and
$B_B=\sum_{i\in B}\ell_{ij}^2$.
The conditional mean batch loss is
\[
 \mu_j=d^{-1}\sum_{B\in\{L,U\}}(h_B/M_B)A_B.
\]
Sampling without replacement gives conditional variance
\[
 v_j=d^{-2}\sum_{B\in\{L,U\}}
 \frac{h_B(M_B-h_B)}{M_B(M_B-1)}(B_B-A_B^2/M_B).
\]
A term with an empty side or deterministic sample is zero.
Therefore
\[
 D=\sum_jw_j\mu_j,\qquad
 V=\sum_jw_jv_j+\sum_jw_j(\mu_j-D)^2.
\]
Tied scores use the same stable rank identities and produce the same endpoint value.

These quantities have an $O(N\log N)$ implementation.
Let $r_i=(Y_i-p_i)/s_i$ and $b_i=-p_i/s_i=r_i-Y_i/s_i$.
For pivot $t$, the endpoint loss is a piecewise linear function with breakpoints $b_i,r_i$.
Below the pivot rank its positive overage contribution is
$(1-\tau)s_i(t-r_i)$.
Above the pivot rank its shortfall contribution is
$\tau\{Y_i-s_i(t-b_i)_+\}$.
Two sorts, by $r_i$ and $b_i$, yield the first and second sums from prefix sums of
$s_i,s_ir_i,s_i^2,s_i^2r_i,s_i^2r_i^2,Y_i,Y_i^2,Y_is_i,Y_is_ib_i$ and their $b_i$ counterparts.
The implemented centered score gauge and a power-of-two response unit improve numerical stability.
An independent dense loss matrix and exact binomial weights check the complete-risk calculation.
Every role-split batch contains all complementary recipients.

\section{Experimental specification}
\label{app:experiments}

\subsection{Reference data}

The six measured tables retain the documented original predictors and target.
\begin{center}
\begin{tabular}{lrrl}
\toprule
Table & Rows & Predictors & Target\\
\midrule
Concrete & 1,030 & 8 & compressive strength\\
Airfoil & 1,503 & 5 & scaled sound pressure\\
Power Plant & 9,568 & 4 & electrical energy output\\
Wine white & 4,898 & 11 & sensory quality\\
Wine red & 1,599 & 11 & sensory quality\\
Superconductivity & 21,263 & 81 & critical temperature\\
\bottomrule
\end{tabular}
\end{center}
Original rows, duplicates, and numeric values are retained.
The registry records the official UCI download URLs, column names, targets, and file identities \citep{uci}.
The experiments study randomly formed audit tasks within each observed reference cohort.

Each simulation has 10,000 rows and 12 initial standard-normal predictors.
The first feature is replaced according to
\[
\begin{array}{ll}
\text{Gaussian} & X_0=.5X_1-.35X_2+.5\epsilon,\\
\text{Heteroscedastic} & X_0=.5X_1+(.15+.6/(1+e^{-X_2}))\epsilon,\\
\text{Heavy-tailed} & X_0=.5X_1+.4t_3,\\
\text{Nonlinear} & X_0=\sin(2X_1)+.3X_2^2+.4\epsilon.
\end{array}
\]
In each case $y=.9X_0+.3X_1-.2X_2+.35\epsilon_y$, with independent draws of the noise terms.

A reference seed permutes the rows.
The first $\min(\lfloor .6n\rfloor,6000)$ fit standardization and a ridge candidate with $\lambda_0=.05$.
The remaining rows form the task cohort.
The hidden feature is the non-intercept column with the largest absolute fitted ridge coefficient.
The first 64 fitting rows supply observed anchors.
The conditional-feature models use the other observed predictors and the observed response.
Their fitting and residual-bank roles remain inside the reference fitting rows.
Task outcomes are computed with the actual withheld feature values.
Each task samples $q$ cohort rows with replacement.
Let $\mathcal A$ contain the 64 observed anchors and let $c_{ij}$ be the number of occurrences of cohort row $j$ in task $i$, so that $\sum_j c_{ij}=q$.
For a completion $v$, its objective is
\[
J_i(\theta;v)=\frac{1}{64+q}\left[
\sum_{a\in\mathcal A}(x_a^\top\theta-y_a)^2+
\sum_{j:c_{ij}>0}c_{ij}(x_j(v_j)^\top\theta-y_j)^2
\right]+.05\|\theta\|_2^2.
\]
Thus every anchor and sampled occurrence has weight $1/(64+q)$, and a repeated cohort identity has total weight $c_{ij}/(64+q)$ after merging.
The intercept is included in $\theta$ and in the ridge penalty.
Its observed outcome is $J_i(\beta;v)-\min_\theta J_i(\theta;v)$ at the actual withheld cohort values.

\paragraph{Working-law and auxiliary quantities.}
\label{app:workingquantities}

Write $n_f=\min(\lfloor .6n\rfloor,6000)$ and $n_I=\lfloor .75n_f\rfloor$.
The first $n_I$ reference fitting rows train the conditional-feature models, using the available standardized covariates and the reference-table regression response as input and the withheld feature as their training target.
The remaining reference fitting rows supply the residual bank used by the auxiliary feature-sampling procedures.

The primary conditional-feature model, denoted $C$, fits nine histogram-gradient-boosted quantile regressors at
\[
\mathcal T=\{.01,.05,.15,.30,.50,.70,.85,.95,.99\}.
\]
Each uses 140 boosting iterations, learning rate $.1$, at most 15 leaves, a minimum of 20 rows per leaf, L2 penalty 1, disabled early stopping, and the reference seed.
Sorting the nine predictions within each row gives $Q^{(C)}_\tau(o_j)$, where $o_j$ denotes the available row input.

The auxiliary model $R$ uses the same $n_I$ fitting rows and a permutation with seed equal to the reference seed plus 1729.
Its location, scale, residual-bank, and selection partitions contain successive proportions $.60,.15,.10,.15$, with cutpoints rounded down.
The location models are ridge regression with penalty 1 and histogram-gradient boosting with 100 iterations, 15 leaves, minimum leaf size 20, and L2 penalty 1.
A 60-iteration, 7-leaf histogram-gradient model, with minimum leaf size 20 and L2 penalty 2, fits $\log(e^2+.01)$ on the scale partition using residuals from the boosted location model.
Its scale prediction is $\operatorname{clip}(\exp(\widehat f(o_j)/2),.02,20)$.
Both boosted models use learning rate $.1$ and disabled early stopping, with the reference seed for location and the reference seed plus one for scale.
Four candidates combine the ridge location with Gaussian or empirical bank residuals, or the boosted location with empirical unscaled or standardized bank residuals.
The Gaussian bank uses its residual mean and standard deviation, floored at $.02$.
The selected candidate minimizes mean pinball loss over the nine levels in $\mathcal T$ on the selection partition, producing $Q^{(R)}_\tau$.

\begin{center}
\small
\begin{tabular}{p{.15\textwidth}p{.76\textwidth}}
\toprule
Quantity & Definition and use\\
\midrule
$m_j$ & $Q^{(C)}_{.50}(o_j)$, the center of the correlated Gaussian working law.\\
$\sigma_j$ & $\max\{(Q^{(C)}_{.85}(o_j)-Q^{(C)}_{.15}(o_j))/2.0729,.02\}$, its row scale.\\
$s_i$ & $\sqrt{t_i^{(C)}t_i^{(R)}}$, the geometric mean of the two auxiliary task standard deviations defined below.\\
$\pi_i$ & $(q_i^{(C)}+q_i^{(R)})/2$, the auxiliary center for standardized-response learners.\\
$\mu_i$ & $\mathbb E[H_i]$ under the correlated Gaussian working law in Equation~\eqref{eq:rowlaw}, computed by the original-input matrix integral.\\
\bottomrule
\end{tabular}
\end{center}

For each auxiliary view $v\in\{C,R\}$, use its median completion $\mathbf m_i^{(v)}$ and row scales
$\sigma_j^{(v)}=\max\{(Q^{(v)}_{.85}(o_j)-Q^{(v)}_{.15}(o_j))/2.0729,.02\}$.
The auxiliary summaries use the diagonal-covariance quadratic Taylor expansion in task-occurrence coordinates, before row-identity merging.
With $D_i^{(v)}=\operatorname{diag}(\sigma_j^{(v)})$, put
\[
c_i^{(v)}=h_i(\mathbf m_i^{(v)}),\qquad
a_i^{(v)}=D_i^{(v)}\nabla h_i(\mathbf m_i^{(v)}),\qquad
B_i^{(v)}=D_i^{(v)}\nabla^2h_i(\mathbf m_i^{(v)})D_i^{(v)}.
\]
Their Gaussian-quadratic cumulants and standard deviations are
\begin{align*}
\kappa_{1,i}^{(v)}&=c_i^{(v)}+\tfrac12\operatorname{tr}B_i^{(v)},&
t_i^{(v)}&=\sqrt{\max\{\|a_i^{(v)}\|_2^2+\tfrac12\|B_i^{(v)}\|_F^2,10^{-16}\}},\\
\kappa_{3,i}^{(v)}&=\operatorname{tr}[(B_i^{(v)})^3]+3(a_i^{(v)})^\top B_i^{(v)}a_i^{(v)}.
\end{align*}
Writing $z_{.95}=\Phi^{-1}(.95)$, the first-order Cornish--Fisher auxiliary forecast is
\[
q_i^{(v)}=\kappa_{1,i}^{(v)}+t_i^{(v)}
\left[z_{.95}+\frac{\kappa_{3,i}^{(v)}}{6(t_i^{(v)})^3}(z_{.95}^2-1)\right].
\]
After undoing source-only target standardization, the public standardized-response view maps its prediction $\widehat r_i$ for $(Y_i-\pi_i)/s_i$ back as $\pi_i+s_i\widehat r_i$.
The functional-kernel update uses $Y_S-\mu_S$ in Equation~\eqref{eq:prediction}.
The residual rule and matched complete-MC control receive the same analytic mean $\mu$.

The fixed auxiliary pilot contains 1152 partial task contexts from reference 4102 for each dataset and task size.
It sets
\[
T=2^{\lceil\log_2\max\{\operatorname{median}_{i\in\mathrm{pilot}}s_i,2^{-40}\}\rceil}.
\]
The implementation uses score scale $s_i^\gamma$. Its equivalent source-normalized gauge is $(s_i/s_{\mathrm{ref}})^\gamma$, with
$s_{\mathrm{ref}}=\exp(640^{-1}\sum_{i\in S}\log s_i)$ and $\gamma\in\{0,1/2,1\}$.
The accompanying reproduction notes identify the implementation entry points.

\subsection{Learning and evaluation roles}

Development uses seeds $4101,4102,4103,4201,4202,4203,4204,4205$.
The original confirmation uses seeds $5101$ through $5120$.
After freezing the shared three-point scale-learning rule, its independent confirmation uses seeds $5301$ through $5320$.
Residual development uses references $4301$ through $4304$, both task sizes, and four independent Monte Carlo streams per condition.
After fixing the residual rule, references $5501$ through $5520$ provide twenty new contexts per dataset.
Each context supplies both task sizes, giving 400 evaluation conditions.
Simulated references redraw their full tables. Measured-table references use new row permutations and tasks within the same six public tables.
The working model has $\eta=.5$, complete first- and second-order coefficients, and the ten ridge values
$10^{-8},10^{-7},\ldots,10^1$.
It uses 640 source outcomes and 512 independent selection outcomes, with scale exponents $\gamma\in\{0,1/2,1\}$.
The residual rule retains the selected degree-two physical ridge and scale exponent, then chooses its weight from $\{0,1/2,1\}$ on the same independent selection pool.
The complete-MC shrinkage control follows the corresponding selection rule with the same 8192 states.
Every candidate in the same condition uses the same tasks.
Selection minimizes the exact complete calibrated loss with $m=256,k=245$ on the independent selection pool.
Development evaluation has 512 tasks, and confirmation evaluation has 2,048 tasks.
For development seed 4102, the auxiliary pilot preprocessing includes the unlabeled evaluation batch through permutation-invariant context statistics.
The predictor and the complete score vector are frozen before calibration roles are assigned, so the fixed-finite-pool coverage argument applies to that condition.
Confirmation task contexts use the previously frozen auxiliary pilot preprocessing.
For split diagnostics, each condition retains the initial split and 64 additional uniform role splits.
Each split has 256 calibration tasks and all complementary test tasks.

For a dataset, the primary reported loss is the arithmetic average of original-unit $D$ across both $q$ values and every listed seed.
The original regret unit is the task-objective unit after reference-table standardization and before the auxiliary response normalizer $T$.
The dataset ratios are then averaged with equal dataset weights.
The minimum over public configurations is computed after this per-dataset aggregation.
Paired bootstrap intervals resample the reference seeds 20,000 times and recompute that public minimum within every resample.
They describe variation over the registered reference mechanism within these tables.
The per-seed and per-$q$ tables accompany the complete comparison.

\subsection{Public learners and hyperparameters}

The classical learners use 175 allowed partial-context features built from the two original task summaries and a frozen pilot transformation.
That transformation uses an auxiliary training context with its allowed task descriptors.
The reproducibility manifest records the exact pilot fields, and the confirmation source bundle contains zero actual-regret pilot members.
The two target views are $Y_i/T$ and $(Y_i-\pi_i)/s_i$, with the auxiliary center $\pi_i$ and scale $s_i$ defined in Appendix~\ref{app:workingquantities}.
Predictions are mapped back to the original response unit before calibration.
Targets are further scaled by their source-only standard deviation.

\begin{center}
\begin{tabular}{lp{.72\textwidth}}
\toprule
Family & Fixed menu and training settings\\
\midrule
Linear QR & $\tau=.95$, source-fitted standardization, L1 penalties $0,10^{-4},10^{-3},10^{-2},10^{-1}$ and HiGHS.\\
Ridge & Source-fitted standardization and penalties $.001,.01,.1,1,10,100,1000$.\\
RF, Extra Trees, QRF & 200 trees, leaf sizes $3,10,30$ crossed with feature fractions $.8,1$. RF and QRF use bootstrap sampling. Extra Trees uses its fully randomized split implementation.\\
Histogram boosting & Quantile loss, 240 iterations, learning rate $.1$, minimum leaf size 12, leaf counts $3,7,15,31$ crossed with L2 penalties $1,10$.\\
LightGBM & Quantile loss, 300 iterations, learning rate $.05$, leaf counts $7,15,31$, minimum child counts $10,30$, L2 penalties $0,1$, and deterministic column-wise execution.\\
\bottomrule
\end{tabular}
\end{center}
These are registered implementations of the respective published families \citep{koenker1978,hoerl1970,breiman2001,geurts2006,meinshausen2006,friedman2001,ke2017,pedregosa2011}.

For each target view and score scale, the public comparison retains three candidate-selection policies.
The first uses source rows 1--512 for fitting and rows 513--640 for pinball selection.
The second selects by complete calibrated loss on those 128 held-out source rows using $m=64,k=62$.
The third selects by complete calibrated loss on the independent 512-task selection pool.
Candidates are refit on all 640 source outcomes.
The kernel candidates use the independent selection pool.
The three selection identities coincide for a fixed one-candidate network, and those repeated identities are retained in the registry.

\subsection{CPCP adaptation}

The archive retains the official CPCP code at commit
\texttt{98ce4fa0c851a9bfdedb609f82d4847f6e666def}, together with its MIT notice \citep{chen2026}.
The adapter preserves its monotonic three-head network, density construction, and fine-tuning procedure.
It uses quantile levels $.93,.95,.97$, 320 source rows for base fitting, 320 for fine-tuning, and 200 epochs per phase.
The optimizer is Adam with learning rate $.002$ and batch size 1024.
The baseline includes vanilla density weighting and Clip+Mix with clipping level five and mixing weight $.5$.
The shared backbone and gap heads are frozen during main-head fine-tuning as in the official routine.
Source-only standardization and the two target views align it with the single-sided task response.
All conditions use deterministic CPU execution and recorded initialization and fine-tuning seeds.

\subsection{Generic task kernels and ablations}

Bag linear and quadratic kernels use weighted means of the allowed normalized row features.
The RBF kernel uses source-standardized context features and the median positive source-pair squared distance as its bandwidth.
The RF-proximity kernel uses common-leaf frequencies in the source-fitted 200-tree RF.
An additional identity-bag kernel uses the exact repeated-row incidence vectors.
Shared-row bag controls use the same
$R_{ij}=.5\mathbf1\{i=j\}+.5\phi_i^\top\phi_j$ row kernel as the functional model, with linear and normalized quadratic variants.
They receive the same operator-mean offset, ridge menu, selection budget, and score scales.
These distribution-regression controls follow established mean-embedding constructions \citep{szabo2016}.

The expanded degree-two registry contains 174 public configurations and 189 total rules.
The original public count comprises 108 classical/network response, scale, and selection configurations, eight initial generic-kernel configurations, eight same-row-kernel bag configurations, and sixteen direct functional ensemble configurations.
The shared scale-learning menu adds eighteen classical/network response configurations, eight generic and shared-row bag configurations, and eight ensemble configurations.
Its fifteen internal rules comprise the degree-two model and its linear, diagonal, independent-row, and kernel-combination ablations under the registered scale policies.
The residual confirmation adds six fixed-weight rules and the two selected-weight rules, giving 197 total registered rules.

\subsection{Shared-state ensemble linear Bayes}

We adapt the ensemble Kalman and linear-Bayes framework of \citet{ernst2015} to the augmented task functionals.
For each common Gaussian row-state draw $v^{(b)}$, compute
\[
 f_i^{(b)}=H_i(P_i v^{(b)}),\qquad
 \widehat C_{ij}=(B-1)^{-1}\sum_{b=1}^B
 (f_i^{(b)}-\widehat\mu_i)(f_j^{(b)}-\widehat\mu_j).
\]
Every task uses the same hidden value whenever a row identity recurs.
The operator retains its anchors, weights, polynomial part, and full rational term.
The row law uses the same $\eta=.5$, centers, scales, and row-feature map as the primary predictor.
Source-mean diagonal normalization and the ten ridge values match the functional-kernel interface.
The standard offset is the ensemble sample mean.
A second offset uses the same supplied operator mean as the matched bag controls, together with the same sample covariance.

Three nested budgets use $B=2048,8192,32768$ and blocks of 2048 draws.
The RNG seed is a fixed SHA256 function of the registered method version, dataset, reference seed, task size, and MC replicate.
Confirmation uses replicate zero.
A separate reference, seed4104, checks Gaussian and Airfoil conditions at both task sizes with replicates zero and one.
The fixed-budget comparison contains twelve rules from the three budgets, two offsets, and two score scales.
Four further rules jointly select budget and ridge on the independent 512-task selection pool.
Each rule then uses the original 256-task calibration role and the same 65 diagnostic role splits.

Production timing records profile reconstruction, sampling and covariance, shared eigensolves, and each offset's ridge fits.
The historical full-projection time identifies the common source of the supplied operator means.
The common three-point scale menu adds eight configurations.
All three budgets and all twenty-four ensemble configurations enter the complete independent scale-learning confirmation panel.

\section{Certified accuracy and integration cost}
\label{app:numericalexperiments}

The original-input matrix panel contains case 0 and case 831 in each of the ten datasets and two task sizes.
The deployment panel contains all 640 sources and query 831 in each of the 20 seed-4101 conditions, giving 12,820 certified functions.
The learned-scale deployment certificates bind the actual independently selected ridges and saved predictions in all twenty development conditions.
Their maximum global coefficient radius is $1.109\times10^{-9}$.
The saved-prediction numerical-error upper bounds range from $2.080\times10^{-11}$ to $3.404\times10^{-6}$, with median $1.113\times10^{-7}$ in original regret units.
The largest ratio of this bound to the absolute saved prediction is $2.100\times10^{-4}$.
The anchored affine-extension squared-risk upper bounds range from $2.708\times10^{-8}$ to $2.077\times10^{-3}$.
Independent 512-bit reconstruction verifies all twenty records, including exact integer global-feature accumulation and the original saved-prediction bindings.

A cross-kernel normalization error in the first certificate reconstruction was corrected before these final records.
The supplementary reproduction notes preserve the original reconstruction, its withdrawal, the correction, and the execution history.

For computation cost, each of the 40 profiles is evaluated at tolerances $10^{-5},10^{-7},10^{-9}$ by three engines.
The common contract requires the mean and Euclidean coefficient radii to be at most the tolerance and the residual-squared interval width to be at most the tolerance times $\max(1,\underline{\E[H^2]})$.
The public baseline uses Arb/Petras integration with the same tilted moments and positive tail bound \citep{johansson2018}.
Its strengthened adapter shares a 32,768-entry tilt cache and validated real-midpoint preconditioning.
All three engines attain the common accuracy requirements on every profile at every tolerance.

\begin{center}
\begin{tabular}{lrrrr}
\toprule
Tolerance & Gamma/Petras cold & Gamma/finite cold & Faster than Petras & Total CPU ratio\\
\midrule
$10^{-5}$ & .06573 & .23134 & 40/40 & .02581\\
$10^{-7}$ & .06458 & .09503 & 40/40 & .02926\\
$10^{-9}$ & .06554 & .10854 & 40/40 & .03226\\
\bottomrule
\end{tabular}
\end{center}
Cold ratios are geometric means over the 40 matched profiles.
The cold timing includes original-input construction, quadrature-rule preparation, quadrature, and certificate extraction in an already imported process.
Engine order is rotated across conditions and quadrature caches are cleared before each calculation.
The CPU column divides total Gamma CPU time by total strengthened-Petras CPU time across all profiles at that tolerance.
Seven Gamma observations across the full tolerance panel fall below the Windows CPU timer's resolution and retain their recorded zero CPU increments.
Per-profile wall times and CPU records are supplied in full.

\begin{figure}[h]
\centering
\includegraphics[width=.55\textwidth]{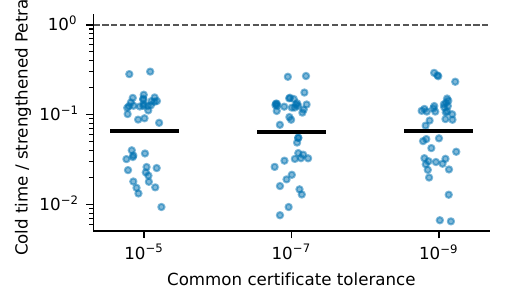}
\caption{Matched original-input certificate cost. Each point is one of 40 profiles and the black bars give geometric means. Both engines meet the same mean, coefficient-radius, and residual-energy contracts.}
\label{fig:cost}
\end{figure}

The current local computation host has an Intel Core Ultra 9 285HX and 64 GB RAM, reserves logical CPUs 0--3, and limits each numerical process to one native thread.
The original confirmation execution uses 48 single-thread workers, and the independent scale-learning and residual confirmations use 96, on a Xeon Gold 5320 server with 256 GB RAM.
The scientific environment versions are NumPy 2.4.3, SciPy 1.17.1, scikit-learn 1.8.0, pandas 2.3.3, LightGBM 4.6.0, and CPU PyTorch 2.15.0.
The supplementary reproduction notes provide module entry points, numerical verification procedures, solver recovery records, and source-reproduction results.

\section{Relation to the earlier regret certificate}

The earlier preprint establishes deterministic completion-relative regret certificates over all continuous completions of an interval table, using residual-energy structure and uniform dimension reduction \citep{li2026}.
It supplies the Schur representation used here.
The present statistical observation model has a collection of scalar task outcomes generated by one shared hidden table state.
Its prediction target is a new task functional, and its algorithm computes a local-to-global projection and residual kernel, original-input matrix and sample certificates, and calibrated task endpoints.
The mathematical guarantees in the present paper concern functional projections, numerical computation, and random task-calibration roles.

\section{Additional evaluation tables}
\label{app:tables}
\IfFileExists{n251_supplement.tex}{\subsection{Residual confirmation uncertainty}
Bootstrap samples resample the twenty reference contexts with task sizes paired and all methods kept together. The public minimum is recomputed in every one of 20000 bootstrap samples.
\begin{center}
\small
\begin{tabular}{lrrrr}
\toprule
Dataset & Original-unit $D$ & Public-min ratio (95\% interval) & Matched-MC ratio (95\% interval) & $V/V_{RF}$\\
\midrule
Gaussian & $0.000288671$ & 0.99273 [0.99010,0.99762] & 0.99590 [0.99267,0.99862] & 0.84927\\
Heteroscedastic & $0.000341602$ & 0.99250 [0.98849,0.99565] & 0.99728 [0.99307,0.99989] & 0.77186\\
Heavy-tailed & $0.00034819$ & 0.99077 [0.98539,0.99724] & 0.98836 [0.98134,0.99387] & 0.58330\\
Nonlinear & $0.000296121$ & 0.99410 [0.99147,0.99692] & 0.99686 [0.99422,0.99911] & 0.85016\\
Concrete & $9.20859e-05$ & 0.98455 [0.97905,0.99035] & 0.99839 [0.99585,1.00086] & 0.08035\\
Airfoil & $0.000312339$ & 0.98815 [0.98016,0.99508] & 0.99568 [0.99154,0.99878] & 0.09946\\
Power Plant & $9.83422e-05$ & 0.99428 [0.99166,0.99706] & 0.99721 [0.99500,0.99897] & 0.65168\\
Wine white & $0.00031186$ & 0.99204 [0.98810,0.99599] & 0.99867 [0.99727,0.99981] & 0.32061\\
Wine red & $0.00035006$ & 0.99353 [0.98681,0.99841] & 0.99610 [0.99356,0.99840] & 0.11773\\
Superconductivity & $6.90933e-05$ & 0.98821 [0.98204,0.99458] & 0.99651 [0.99211,1.00033] & 0.48456\\
\bottomrule
\end{tabular}
\end{center}
\subsection{Degree-two ablation on the same confirmation panel}
\label{app:k2ablation}
The following comparison uses the frozen residual rule and its degree-two component on the same references 5501--5520, both task sizes, and all 400 confirmation conditions.
Each dataset first averages original-unit complete risk across references and task sizes. The paired intervals resample reference contexts while retaining both task sizes together.
The mean dataset ratio is 0.99767734, with a paired interval of [0.99632446,0.99880885].
Point estimates are lower on 7 of ten datasets, and the interval upper bound is below one on 5 of ten.
\begin{center}
\small
\begin{tabular}{lrr}
\toprule
Dataset & Residual/$K_2$ & 95\% paired interval\\
\midrule
Gaussian & 0.9998769 & [0.9997595,0.9999675]\\
Heteroscedastic & 0.9997732 & [0.9995992,0.9999187]\\
Heavy-tailed & 0.9824890 & [0.9713896,0.9918890]\\
Nonlinear & 0.9999662 & [0.9998860,1.0000345]\\
Concrete & 1.0007818 & [0.9987406,1.0025781]\\
Airfoil & 0.9953142 & [0.9900476,0.9993714]\\
Power Plant & 0.9997120 & [0.9994480,0.9999315]\\
Wine white & 1.0001700 & [0.9999758,1.0004059]\\
Wine red & 0.9986850 & [0.9962588,1.0007728]\\
Superconductivity & 1.0000053 & [0.9999148,1.0000909]\\
\bottomrule
\end{tabular}
\end{center}
\subsection{Residual confirmation by task size}
\begin{center}
\small
\begin{tabular}{lrrrr}
\toprule
Dataset & Public ratio $q=8$ & Public ratio $q=32$ & Matched ratio $q=8$ & Matched ratio $q=32$\\
\midrule
Gaussian & 0.99133 & 0.99684 & 0.99711 & 0.99502\\
Heteroscedastic & 0.99463 & 0.99097 & 0.99815 & 0.99664\\
Heavy-tailed & 0.99135 & 0.99036 & 0.98657 & 0.98970\\
Nonlinear & 0.99554 & 0.99305 & 0.99728 & 0.99654\\
Concrete & 0.98295 & 0.98530 & 0.99675 & 0.99915\\
Airfoil & 0.98649 & 0.98889 & 0.99849 & 0.99445\\
Power Plant & 0.99652 & 0.99271 & 0.99782 & 0.99679\\
Wine white & 0.99247 & 0.99175 & 0.99995 & 0.99780\\
Wine red & 0.99022 & 0.99523 & 0.99473 & 0.99680\\
Superconductivity & 0.98935 & 0.99020 & 0.99587 & 0.99702\\
\bottomrule
\end{tabular}
\end{center}
}{}
\subsection{Earlier degree-two confirmation}
The following tables retain the degree-two rule's independent confirmation on references 5301--5320.
\IfFileExists{supplementary_results.tex}{\subsection{Dataset-level uncertainty}
Intervals resample reference seeds jointly across methods and recompute the retrospective public minimum.
\begin{center}
\small
\begin{tabular}{lrrrr}
\toprule
Dataset & $D$ & Public-min ratio & 95\% interval & $V/V_{\mathrm{RF}}$\\
\midrule
Gaussian & 0.0002799 & 0.9936 & [0.9903, 0.9970] & 0.8626\\
Heteroscedastic & 0.00034303 & 0.9959 & [0.9936, 0.9983] & 0.8143\\
Heavy-tailed & 0.0003667 & 0.9982 & [0.9923, 1.0102] & 0.8008\\
Nonlinear & 0.00029564 & 0.9952 & [0.9921, 0.9986] & 0.7868\\
Concrete & 8.7718e-05 & 0.9787 & [0.9721, 0.9855] & 0.0649\\
Airfoil & 0.00032099 & 1.0179 & [1.0010, 1.0350] & 0.2031\\
Power Plant & 9.3969e-05 & 0.9962 & [0.9937, 1.0003] & 0.7348\\
Wine white & 0.00031644 & 0.9878 & [0.9818, 0.9935] & 0.2769\\
Wine red & 0.00031296 & 0.9918 & [0.9883, 0.9957] & 0.1762\\
Superconductivity & 6.9474e-05 & 0.9952 & [0.9912, 1.0021] & 0.7598\\
\bottomrule
\end{tabular}
\end{center}
\subsection{Task-size strata}
\begin{center}
\small
\begin{tabular}{lrrrr}
\toprule
Dataset & $q=8$ ratio & $q=32$ ratio & $D_{q=8}$ & $D_{q=32}$\\
\midrule
Gaussian & 0.98961 & 0.99652 & 0.00023473 & 0.00032508\\
Heteroscedastic & 0.99634 & 0.99553 & 0.00028843 & 0.00039762\\
Heavy-tailed & 0.98861 & 1.0054 & 0.00031184 & 0.00042156\\
Nonlinear & 0.99705 & 0.9938 & 0.0002493 & 0.00034197\\
Concrete & 0.97353 & 0.98125 & 5.6402e-05 & 0.00011904\\
Airfoil & 1.0267 & 1.0138 & 0.00020688 & 0.00043511\\
Power Plant & 0.99343 & 0.9983 & 7.8108e-05 & 0.00010983\\
Wine white & 0.99132 & 0.98631 & 0.00026258 & 0.0003703\\
Wine red & 0.99317 & 0.99167 & 0.00021122 & 0.0004147\\
Superconductivity & 0.99945 & 0.99497 & 6.0579e-05 & 7.8369e-05\\
\bottomrule
\end{tabular}
\end{center}
\subsection{Named public families}
Entries divide the earlier degree-two rule's dataset loss on references 5301--5320 by the minimum dataset loss among the registered configurations of each named family. These are retrospective within-family minima for that earlier confirmation.
\begin{center}
\small
\begin{tabular}{lrrrrrrr}
\toprule
Dataset & QR & Ridge & RF & ET & QRF & HGB & LGBM\\
\midrule
Gaussian & 0.914 & 0.913 & 0.914 & 0.914 & 0.912 & 0.903 & 0.903\\
Heteroscedastic & 0.894 & 0.899 & 0.896 & 0.896 & 0.896 & 0.888 & 0.888\\
Heavy-tailed & 0.931 & 0.961 & 0.963 & 0.960 & 0.958 & 0.939 & 0.940\\
Nonlinear & 0.884 & 0.894 & 0.894 & 0.891 & 0.890 & 0.884 & 0.885\\
Concrete & 0.254 & 0.259 & 0.264 & 0.264 & 0.264 & 0.258 & 0.259\\
Airfoil & 0.457 & 0.451 & 0.460 & 0.462 & 0.471 & 0.460 & 0.463\\
Power Plant & 0.872 & 0.880 & 0.879 & 0.880 & 0.889 & 0.876 & 0.879\\
Wine white & 0.570 & 0.581 & 0.594 & 0.602 & 0.596 & 0.586 & 0.585\\
Wine red & 0.409 & 0.301 & 0.430 & 0.432 & 0.434 & 0.427 & 0.429\\
Superconductivity & 0.781 & 0.796 & 0.842 & 0.848 & 0.872 & 0.848 & 0.853\\
\bottomrule
\end{tabular}
\end{center}
\subsection{Networks and generic task kernels}
The same retrospective within-family minimum is used in this table.
\begin{center}
\small
\begin{tabular}{lrrrrr}
\toprule
Dataset & CPCP & Clip+Mix & Bag kernels & RBF & RF proximity\\
\midrule
Gaussian & 0.864 & 0.865 & 0.955 & 0.906 & 0.902\\
Heteroscedastic & 0.856 & 0.857 & 0.944 & 0.890 & 0.884\\
Heavy-tailed & 0.917 & 0.912 & 0.796 & 0.915 & 0.866\\
Nonlinear & 0.849 & 0.852 & 0.920 & 0.887 & 0.880\\
Concrete & 0.250 & 0.251 & 0.412 & 0.262 & 0.263\\
Airfoil & 0.440 & 0.443 & 0.590 & 0.458 & 0.461\\
Power Plant & 0.846 & 0.850 & 0.918 & 0.873 & 0.867\\
Wine white & 0.563 & 0.567 & 0.656 & 0.598 & 0.597\\
Wine red & 0.403 & 0.409 & 0.579 & 0.417 & 0.428\\
Superconductivity & 0.773 & 0.785 & 0.672 & 0.778 & 0.792\\
\bottomrule
\end{tabular}
\end{center}
The CSV archive provides configuration identities, per-reference losses, loss components, and role-split metrics.
}{The complete registered evaluation tables are generated after the independent confirmation audit.}
\IfFileExists{ensemble_supplement.tex}{\subsection{Direct ensemble controls}
Entries divide the functional-kernel dataset loss by the minimum across the two offsets and three score-scale rules at each listed budget.
The final column uses the six rules that select budget and ridge on the independent selection pool.
\begin{center}
\small
\begin{tabular}{lrrrr}
\toprule
Dataset & 2,048 & 8,192 & 32,768 & Selected budget\\
\midrule
Gaussian & 0.9276 & 0.9728 & 0.9936 & 0.9880\\
Heteroscedastic & 0.9304 & 0.9742 & 0.9959 & 0.9935\\
Heavy-tailed & 0.9067 & 0.9725 & 0.9982 & 0.9963\\
Nonlinear & 0.9284 & 0.9758 & 0.9952 & 0.9901\\
Concrete & 0.8685 & 0.9545 & 0.9787 & 0.9759\\
Airfoil & 0.9220 & 0.9981 & 1.0179 & 1.0159\\
Power Plant & 0.9289 & 0.9761 & 0.9962 & 0.9913\\
Wine white & 0.8832 & 0.9613 & 0.9878 & 0.9811\\
Wine red & 0.9133 & 0.9756 & 0.9918 & 0.9903\\
Superconductivity & 0.9426 & 0.9785 & 0.9952 & 0.9929\\
\bottomrule
\end{tabular}
\end{center}
All ensembles use the same complete task functions, shared row identities, Gaussian row law, source labels, selection labels, ridge menu, and task-calibration roles as the functional model.
Sampling is streamed in blocks of 2,048 draws. Each state contributes to every task before its block is released.
The supplied operator means are shared with the matched bag controls.
The cost archive records profile reconstruction, sampling and covariance, eigensolves, each offset's ridge fits, independent audits, synthetic-function calls, and peak worker memory.
The archived full-projection time attached to the supplied means identifies their common construction source.
A separate non-confirmation reference supplies eight two-replicate qualification conditions and 64 paired MC-risk comparisons.
}{}
\IfFileExists{scale_supplement.tex}{\subsection{Scale learning and the original confirmation}
The original common-scale rule used references 5101--5120 and led all 140 public configurations on seven of ten datasets.
The three-point scale menu and joint selection rule were fixed before the independent references 5301--5320.
The expanded public comparison includes all original configurations and 34 matched scale-learning configurations.
The following table compares the two functional rules on the same new references and records all frozen scale choices.
\begin{center}
\small
\begin{tabular}{lrrrr}
\toprule
Dataset & Learned/common $D$ & $\gamma=0$ & $\gamma=1/2$ & $\gamma=1$\\
\midrule
Gaussian & 0.9967 & 10 & 16 & 14\\
Heteroscedastic & 0.9968 & 10 & 19 & 11\\
Heavy-tailed & 0.9719 & 19 & 20 & 1\\
Nonlinear & 0.9919 & 7 & 23 & 10\\
Concrete & 0.9580 & 16 & 21 & 3\\
Airfoil & 0.9622 & 5 & 33 & 2\\
Power Plant & 1.0020 & 2 & 7 & 31\\
Wine white & 0.9893 & 4 & 27 & 9\\
Wine red & 0.9767 & 10 & 24 & 6\\
Superconductivity & 1.0012 & 0 & 19 & 21\\
\bottomrule
\end{tabular}
\end{center}
\begin{figure}[ht]
\centering
\includegraphics[width=.9\textwidth]{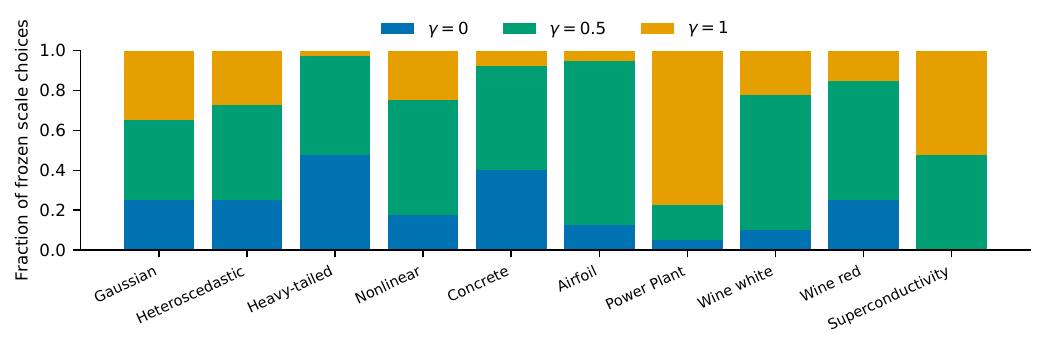}
\caption{Frozen score-scale choices for the earlier degree-two rule on references 5301--5320 and both task sizes. Each choice uses the independent selection pool.}
\label{fig:scalechoices}
\end{figure}
}{}
\section{Independent score-scale learning}
\label{app:scalelearning}

Let $s_i>0$ be the reference-fitted task scale, and let $s_{\mathrm{ref}}$ be its geometric mean over source tasks.
For each fixed point-prediction candidate $a$, consider
\[
 s_{\gamma,i}=(s_i/s_{\mathrm{ref}})^\gamma,
 \qquad \gamma\in[0,1].
\]
Multiplying every task scale by one common positive constant leaves Equation~\eqref{eq:upper} unchanged.
The implementation therefore stores the equivalent gauge $s_i^\gamma$.
The finite menu $\Gamma=\{0,1/2,1\}$ includes the unit-scale and common-scale rules.
On 512 independent selection tasks, the algorithm jointly minimizes complete calibrated loss over the point-prediction candidates and $\Gamma$.
Equal risks prefer $\gamma$ closest to one and then the smaller original candidate index.
Every public predictor receives this same menu and selection budget.
The selected predictor and scale are fixed before calibration roles are assigned.

\begin{proposition}[Scale-grid approximation on a fixed pool]
\label{prop:scalegrid}
Fix a point predictor $p$, labels $Y$, positive scales $s$, and $1\le k\le m<N$.
Put $e_j=Y_j-p_j$ and $t_{ij}=s_i/s_j$.
The complete risk as a function of $\gamma\in[0,1]$ is Lipschitz with constant
\[
 L_N=\frac{.95}{N}\sum_{i=1}^N
 \max_{j\ne i}|e_j|\max(1,t_{ij})|\log t_{ij}|.
\]
If a grid contains both endpoints and has maximum gap $\Delta$, then
\[
 0\le\min_{\gamma\in\Gamma}D_N(\gamma)
 -\min_{\gamma\in[0,1]}D_N(\gamma)
 \le L_N\Delta/2.
\]
\end{proposition}

\begin{proof}
For a fixed recipient $i$ and calibration subset $\mathcal C$, write the endpoint as
\[
 U_{i,\mathcal C}(\gamma)=\max\left\{0,p_i+
 \operatorname{kth}_{j\in\mathcal C}[e_jt_{ij}^{\gamma}]\right\}.
\]
Each curve inside the order statistic has derivative bounded in absolute value by
\[
 |e_j|\max(1,t_{ij})|\log t_{ij}|.
\]
An order statistic is nonexpansive in the input maximum norm, as is clipping at zero.
Pinball loss is $.95$-Lipschitz in its endpoint.
The recipient has marginal weight $1/N$ in complete risk, giving the stated constant after averaging.
Continuity covers sorting ties and clipping transitions.
Every continuous minimizer lies within $\Delta/2$ of a grid point, proving the grid bound.
\end{proof}

For joint selection, let $d_a=\min_{\gamma\in\Gamma}D_{N,a}(\gamma)$ and
$\varepsilon_a=L_{N,a}/4$ for each point-prediction candidate $a$.
The complete continuous-family optimum lies in
\[
 \left[\max\{0,\min_a(d_a-\varepsilon_a)\},\ \min_a d_a\right].
\]
This calculation concerns the fixed selection pool.
The independent confirmation evaluates the resulting frozen learning rule.
The archive records point-arithmetic diagnostics for every candidate's Lipschitz expression and grid minimum.

\section{Multiscale subsampling of complete task risk}
\label{app:subsampling}

The following results apply classical U-statistic symmetrization and finite-population orthogonal decomposition to the complete calibrated loss \citep{hoeffding1948,bloznelis2001}.
Fix the recorded labels, predictions, positive scales, and candidate identity.
Set $h=m+1$ and define, for each $h$-element set $A$,
\[
 g(A)=\frac1h\sum_{i\in A}
 \loss\left(Y_i,\max\{0,p_i+s_i S_{A\setminus\{i\},(k)}\}\right).
\]
For $B\subseteq[N]$ of size $b\ge h$, write
\[
 D_B=\binom bh^{-1}\sum_{\substack{A\subseteq B\\|A|=h}}g(A).
\]

\begin{proposition}[Exact subsampling mean]
\label{prop:subsamplemean}
Equation~\eqref{eq:risk} equals
\[
 D_N=\binom Nh^{-1}\sum_{|A|=h}g(A).
\]
If $B$ is a uniform subset of size $b$ and $m,k$ remain fixed, then
\[
 \E_B D_B=D_N.
\]
The equality also holds for the paired difference of two fixed candidates.
\end{proposition}

\begin{proof}
The identity $\binom Nm(N-m)=h\binom Nh$ matches every calibration set and complementary recipient to one $h$-set and its designated recipient.
For any fixed $h$-set $A$, its inclusion probability in $B$ is
$\binom{N-h}{b-h}/\binom Nb$.
Multiplying this probability by $1/\binom bh$ gives $1/\binom Nh$.
Linearity gives the paired-difference identity.
\end{proof}

For a finite family of candidates, selecting a new minimizer within each subsample gives
$\E_B\min_a D_B(a)\le\min_a D_N(a)$.
The diagnostic experiment keeps the candidate fixed throughout the subset draws.

Assume $h<N$ and put $d=\min(h,N-h)$.
The finite-population Hoeffding decomposition gives
\[
 g(A)=D_N+\sum_{s=1}^{d}G_s(A),\qquad
 G_s(A)=\sum_{\substack{S\subseteq A\\|S|=s}}\psi_s(S),
\]
where the component functions satisfy
\[
 \sum_{j\notin T}\psi_s(T\cup\{j\})=0
 \qquad\text{for every }|T|=s-1.
\]
Define the nonnegative $h$-slice variance components
\[
 v_s=\frac{\binom{N-2s}{h-s}}{\binom Nh}
 \sum_{|S|=s}\psi_s(S)^2.
\]
The orthogonality gives $\operatorname{Var}_{|A|=h}(g(A))=\sum_{s=1}^d v_s$.

\begin{proposition}[Exact multiscale variance]
\label{prop:subsamplevariance}
For a uniform $b$-subset $B$, with $h\le b\le N$,
\[
 \operatorname{Var}_B(D_B)=\sum_{s=1}^{d}\omega_s(b)v_s,
 \qquad
 \omega_s(b)=\frac{\binom hs\binom{N-b}s}
 {\binom bs\binom{N-h}s}.
\]
Here $\binom ar=0$ for $r>a$.
\end{proposition}

\begin{proof}
Averaging the decomposition of $g$ over $h$-sets in $B$ yields
\[
 D_B-D_N=\sum_{s=1}^d\frac{\binom hs}{\binom bs}G_s(B).
\]
The degeneracy identities imply orthogonality across orders and
\[
 \E_B[G_s(B)^2]=\frac{\binom{N-2s}{b-s}}{\binom Nb}
 \sum_{|S|=s}\psi_s(S)^2.
\]
One way to obtain these identities is to associate the multilinear polynomial
$F_t(x)=\sum_{|T|=t}\psi_t(T)\prod_{j\in T}x_j$ with each order.
Degeneracy gives $\sum_j\partial_jF_t=0$, hence $F_t(x+a\mathbf1)=F_t(x)$.
For a fixed set $S$, this implies
\[
 \E[G_t(B)\mid S\subseteq B]
 =\frac{(N-b)_t}{(N-|S|)_t}G_t(S),
\]
where $(a)_t$ denotes a falling factorial.
Expanding $G_s(B)$ and using $G_t(S)=0$ for $|S|<t$ proves cross-order orthogonality.
Using $|S|=t$ gives the displayed second moment.
Substitution and cancellation of the combinatorial factors give $\omega_s(b)$.
\end{proof}

Let $D_{-i}$ denote the risk after deleting task $i$, and define
\[
 J=\frac{N-1}{N}\sum_{i=1}^N(D_{-i}-D_N)^2.
\]
Proposition~\ref{prop:subsamplemean} gives $N^{-1}\sum_iD_{-i}=D_N$.

\begin{corollary}[Identified first-order variation and remainder]
\label{cor:subsampling}
The exact first-order component satisfies $v_1=(N-h)J/h$.
For $h\le b\le N-1$, define
$T_b=b\operatorname{Var}_B(D_B)/(N-b)$ and $R_b=T_b-J$.
Then
\[
 T_b\ge J,\qquad T_c\le T_b\quad(b\le c\le N-1),\qquad T_{N-1}=J.
\]
For $h\le b\le c\le N-1$ and $b\le N-2$,
\[
 0\le R_c\le
 \frac{(b-1)(N-c-1)}{(c-1)(N-b-1)}R_b.
\]
\end{corollary}

\begin{proof}
At $b=N-1$, every order above one vanishes and
\[
 D_{-i}-D_N=-h\psi_1(\{i\})/(N-1).
\]
Substitution in $J$ gives the identity for $v_1$.
For every visible order define
\[
 r_s(b)=\frac{\omega_s(b)}{\omega_1(b)}
 =\prod_{\ell=1}^{s-1}
 \frac{(h-\ell)(N-b-\ell)}{(b-\ell)(N-h-\ell)},
\]
and set $r_s(b)=0$ when $s>N-b$.
Then $T_b=J+h\sum_{s=2}^d r_s(b)v_s/(N-h)$.
All factors decrease as $b$ increases and all variance components are nonnegative.
For $s\ge2$ visible at $c$, the ratio $r_s(c)/r_s(b)$ is a product of factors at most one.
Bounding that product by its first factor gives the stated remainder contraction after summing.
\end{proof}

The first-order approximation to the conditional subset variance is $(N-b)J/b$, with exact remainder $(N-b)R_b/b$.
These statements also apply to a fixed paired risk difference by replacing $g$ with the difference kernel.
The random object in these formulas is a uniform subset of the fixed labeled task pool.
For the development pool, $N=512,h=257$, so the highest orthogonal order is $255$.

The empirical diagnostic uses the known complete risk to form
\[
 Q_{b,r}=(D_{B_{b,r}}-D_N)^2,\qquad
 \widehat T_b=\frac b{N-b}\frac1R\sum_{r=1}^R Q_{b,r}.
\]
This is an unbiased Monte Carlo estimate of $T_b$.
Shared nested subsets give paired Monte Carlo standard errors for the monotonicity and remainder comparisons.
The archive records each condition's estimates and Monte Carlo standard errors.

\subsection{Paired multiscale subsampling experiment}
The development reference 4101 supplies all ten datasets and both task sizes, with $N=512$, $m=256$, and $k=245$.
For each condition, 256 random permutations form nested subsets of sizes 358, 410, and 461.
The frozen functional predictor and the fixed RF reference share every subset.
The experiment records 30,720 complete subset risks and 20,480 delete-one calculations.
The first subset at each size is independently checked with the dense risk calculation for both methods, giving 120 checks.
Exact rational enumeration at $N=7$ and $N=10$, with $m=3,k=3$, reconstructs every subset-size variance from its nonnegative orthogonal components and verifies the mean identity, paired differences, jackknife lower bound, and remainder contraction.
Across the 120 condition-method-size means, the largest absolute Monte Carlo standardized discrepancy from $D_N$ is 2.615.
For the 60 paired differences it is 1.850.
\begin{figure}[ht]
\centering
\includegraphics[width=.95\textwidth]{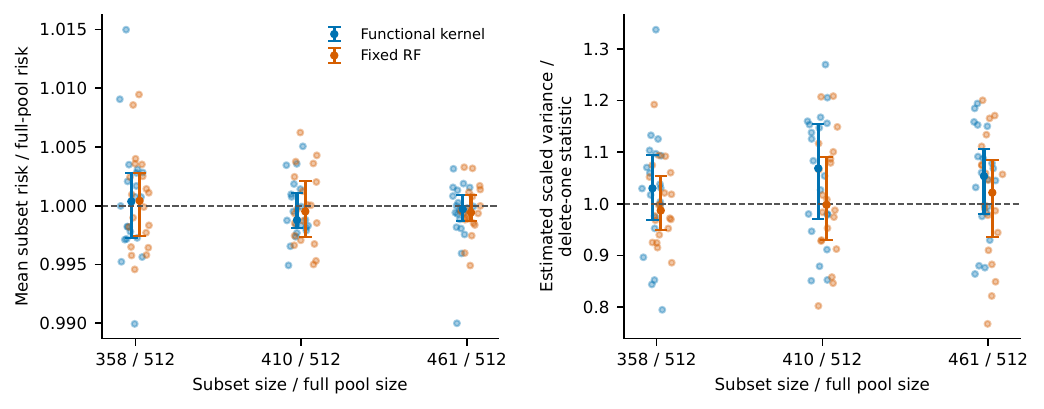}
\caption{Multiscale subsampling on all twenty development conditions. Left, the mean subset risk divided by the complete full-pool risk. Right, $b\widehat{\operatorname{Var}}_B(D_B)/(N-b)$ divided by the delete-one jackknife statistic. Points show conditions and bars show medians and interquartile ranges. Each estimated subset variance uses 256 nested paired subsamples.}
\label{fig:subsampling}
\end{figure}

\section{Orthogonal residual completion and complete-risk perturbation}
\label{app:residualcompletion}

Fix the task operators, row contexts, and Gaussian working law.
Write the complete functional vector as $H=\mu+F+r$, where $F$ is its projection onto the common first- and second-order Gaussian chaos.
Theorem~\ref{thm:kernel} gives $\E[Fr^\top]=0$ and $\operatorname{Cov}(H)=K_2+K_r$, with $K_2=\operatorname{Cov}(F)$ and $K_r=\operatorname{Cov}(r)$.

\begin{proposition}[Shared residual completion]
Let $M\geq2$ independent common states generate residual vectors $r^{(b)}$, and let $\bar r=M^{-1}\sum_b r^{(b)}$.
Then
\[
\widehat K_M=K_2+\frac1{M-1}\sum_{b=1}^M
(r^{(b)}-\bar r)(r^{(b)}-\bar r)^\top
\]
is positive semidefinite and unbiased for $\operatorname{Cov}(H)$.
It converges almost surely when the task functions have finite second moments.
Using the same state-index and row-identity random map for future queries preserves every existing covariance block.
For a fixed finite task collection with $\E\|r\|^4<\infty$,
\begin{align*}
\E\|\widehat K_M-\operatorname{Cov}(H)\|_F^2
=\frac1M\Bigl[&\E\|r\|^4-\frac{M-2}{M-1}\|K_r\|_F^2\\
&+\frac{(\tr K_r)^2}{M-1}\Bigr].
\end{align*}
\end{proposition}

\begin{proof}
The residual term is a centered Gram matrix.
Its expectation is $K_r$, and entrywise laws of large numbers yield convergence.
The common feature map embeds every enlarged finite task set into the same Gaussian and residual-sample coordinates.
Summing the standard sample-covariance element variances gives the displayed fourth-moment expression.
\end{proof}

The residual completion estimates covariance under the fixed working law.
Its finite-budget prediction performance is assessed by complete calibrated loss.
The registered development experiments retain each independent Monte Carlo stream as an algorithm repetition.

\begin{proposition}[Numerical projection error]
Let $\Gamma$ denote the exact common coefficient matrix, and let the stored polynomial use $\widetilde\Gamma=\Gamma+E$.
If its exact Gram is $\widetilde K_2$ and $\|E\|_F\leq\epsilon_\Gamma$, then
\begin{align*}
\widetilde K_2+\operatorname{Cov}(H-\widetilde\mu-\widetilde F)
-\operatorname{Cov}(H)
&=\widetilde\Gamma E^\top+E\widetilde\Gamma^\top,\\
\bigl\|\widetilde K_2+\operatorname{Cov}(H-\widetilde\mu-\widetilde F)
-\operatorname{Cov}(H)\bigr\|_{\rm op}
&\leq2\|\widetilde\Gamma\|_{\rm op}\epsilon_\Gamma.
\end{align*}
\end{proposition}

\begin{proof}
Centering removes the constant mean error.
The centered implemented residual is $r-e$, where $e$ has coefficient matrix $E$ and lies in the common degree-two chaos.
Thus $\E[re^\top]=0$.
Expanding the polynomial Gram and residual covariance gives the identity.
The operator-norm bound follows from submultiplicativity and $\|E\|_{\rm op}\leq\|E\|_F$.
\end{proof}

Floating-point Gram assembly and complete-operator evaluation contribute their separately enclosed numerical errors.
The coefficient radii from the original-input matrix construction control $\epsilon_\Gamma$.

\begin{proposition}[Fixed-weight covariance error]
For a fixed weight $\omega\in[0,1]$ chosen independently of the Monte Carlo states, let
$V_M=\E\|\widehat K_r-K_r\|_F^2$ and $A_r=\|K_r\|_F^2$.
Then
\[
\E\|K_2+\omega\widehat K_r-\operatorname{Cov}(H)\|_F^2
=(1-\omega)^2A_r+\omega^2V_M.
\]
When $A_r+V_M>0$, the minimizing fixed weight is $A_r/(A_r+V_M)$.
\end{proposition}

\begin{proof}
The centered sample covariance is unbiased for $K_r$, so the cross term between its zero-mean estimation error and the fixed shrinkage bias vanishes.
Expanding the square and differentiating the resulting quadratic gives the claims.
\end{proof}

The experiment selects its weight by complete calibrated loss on an independent task pool.
The displayed covariance error quantifies the fixed-weight bias and variance mechanism.

For a fixed physical-unit ridge $\tau>0$, let $A=K_{2,ss}+\tau I$, $R=K_{r,ss}$, $k=K_{2,s*}$, $t=K_{r,s*}$, and $y=Y_s-\mu_s$.
The resolvent identity gives
\[
p_H-p_2=(t-RA^{-1}k)^\top(A+R)^{-1}y.
\]
This equality identifies the source--query residual direction that changes the prediction.
The mechanism experiment evaluates it alongside a source-diagonal loading comparison.

\begin{theorem}[Complete-risk numerical perturbation]
\label{thm:riskperturbation}
Fix nonnegative labels $Y_i$, positive scales $s_i$, a pool size $N$, and $1\leq k\leq m<N$.
Suppose $|\widetilde p_i-p_i|\leq\epsilon_i$ and write $a_i=\epsilon_i/s_i$.
Order the $a_i$ increasingly, breaking ties by a fixed task index, and order the scales correspondingly.
Let
\[
v_j=\frac{\binom{j-1}{m-1}}{\binom Nm},\qquad
L=\max(\vartheta,1-\vartheta),
\]
where $\vartheta$ is the pinball level.
Then
\begin{align*}
|D(\widetilde p)-D(p)|
\leq L\Biggl[\bar\epsilon+
\sum_{j=m}^N\frac{v_ja_{(j)}}{N-m}
\biggl\{\sum_{i>j}s_{(i)}
+\frac{j-m}{j-1}\sum_{i<j}s_{(i)}\biggr\}\Biggr].
\end{align*}
The second term in braces is zero when $j=m=1$.
The bound is computable in $O(N\log N)$ real-arithmetic time and $O(N)$ storage, and is at most
$L(\bar\epsilon+\bar s\max_i a_i)$.
\end{theorem}

\begin{proof}
For each calibration set $C$, the normalized scores change by at most $\max_{j\in C}a_j$.
The order statistic and clipping maps are nonexpansive, so the endpoint for $i\notin C$ changes by at most
$\epsilon_i+s_i\max_{j\in C}a_j$.
The pinball loss is $L$-Lipschitz in this endpoint.
In complete averaging, each task has marginal recipient weight $1/N$, giving $\bar\epsilon$.
The largest sorted position in a uniform calibration set is $j$ with probability $v_j$.
Conditional on this event, all positions above $j$ are recipients and a uniform $j-m$ of the $j-1$ lower positions are recipients.
Taking their scale sum and dividing by $N-m$ gives the displayed expression.
Sorted prefix and suffix sums give the complexity.
Replacing each calibration maximum by the global maximum gives the simpler bound.
\end{proof}

If a fixed finite candidate family has numerical risk error at most $\epsilon_D$ for every candidate on one selection pool, its numerical minimizer has exact selection-pool risk at most $2\epsilon_D$ above the exact candidate minimum.
This follows by applying the uniform bound at the two minimizing candidates.
The error radius therefore connects original-input coefficient enclosures to the complete-risk selection computation.

An Arb implementation orders the binary64 ratios by exact rational comparison and encloses the positive weighted sum.
Exact rational enumeration verifies the calibration expectation on 18 configurations with $N\in\{5,7,10\}$, including tied ratios, $m=1$, and $m=N-1$.

\paragraph{A complete-pool numerical certificate.}
For Airfoil reference 4101 with $q=8$, we enclose all 512 evaluation-query functions and combine them with the 640 previously audited source-function enclosures.
The certificate uses the actual independently selected $\lambda=.01$, $\gamma=1/2$, source coefficients, and saved predictions.
Exact score ordering and positive Arb pivot sums also enclose the complete risk of the stored predictions, including its own evaluation arithmetic.
The stored-predictor risk is $1.7302025840184\times10^{-4}$.
Theorem~\ref{thm:riskperturbation} gives a numerical-risk error radius of at most $8.273\times10^{-7}$, yielding an exact-working-kernel risk interval contained in $[1.7219,1.7385]\times10^{-4}$.
The sorted calibration term reduces the global-maximum radius by 17.61\% on this pool.
The recipient-conditioned summation and high-precision reconstruction procedures are specified in the reproduction notes.

\section{Numerical certification of the finite-state residual predictor}
\label{app:residualnumerics}

Fix the original binary64 operator inputs, their exact real interpretation, and the stored primitive normal draws.
The exact Gaussian map of these draws defines each sample state.
Source observations, scales, and the selected ridge, scale exponent, and residual weight are fixed.
The target predictor uses the exact low-order coefficients and the exact centered residual features at these finite states.
Its kernel is the PSD matrix in Equation~\eqref{eq:residualkernel}.

\subsection{Sample enclosures and centering}

Arb encloses the original Schur coefficients once per task.
Outward binary64 interval operations then enclose all quadratic forms and the rational quotient at the fixed states, using the positive denominator floor.
The exact shared Gaussian map is included in these intervals.
Comparing the interval with the production complete-function value gives $e_{H,ib}$.

Let $\zeta_i\in\mathbb R^{d_i}$ contain the distinct private-row draws and the shared draws used by task $i$'s lifted polynomial.
The normalized first- and second-order Hermite vector contains $\zeta_{ij}$, $(\zeta_{ij}^2-1)/\sqrt2$, and $\zeta_{ij}\zeta_{ik}$ for $j<k$.
It excludes the constant feature, and satisfies
\[
\|\Psi_2(\zeta_i)\|_2^2=\frac{\|\zeta_i\|_2^4+d_i}{2}.
\]
If the global coefficient radius is $e_i$, its pointwise contribution is bounded by $e_i\|\Psi_2(\zeta_i)\|_2$.
We also enclose the stored lifted polynomial directly and compare it with the production local evaluation, including the state-generation and lifting arithmetic.
The resulting bound is $e_{F,ib}$.
The two subtraction operations in the production residual contribute their floating-point error bound $e_{\mathrm{sub},ib}$.
Thus the residual evaluation error vector is bounded entrywise by
\[
b_{ib}=e_{H,ib}+e_{F,ib}+e_{\mathrm{sub},ib}.
\]

Let $J=I-\mathbf1\mathbf1^\top/M$.
Centering removes each taskwise constant mean offset.
Since $\|J\|_{\mathrm{op}}=1$, the normalized centered-row error is at most
\[
e_i^{(r)}=\frac{\|b_i\|_2}{\sqrt{M-1}}+e_{\mathrm{center},i},
\]
where $e_{\mathrm{center},i}$ encloses the actual centered-feature subtraction and its remaining row sum.
The implementation uses outward Euclidean norm bounds and standard floating-point sum bounds.

\subsection{Separate Gram blocks}

Let $S_2,S_r$ bound the source Frobenius norms of the stored low-order and residual feature blocks.
Let $E_2,E_r$ bound the corresponding Frobenius errors.
For source assembly and the query cross vector, use
\begin{align*}
d_K={}&2S_2E_2+E_2^2+\omega(2S_rE_r+E_r^2)+A_K,\\
d_{k_i}={}&S_2e_i^{(2)}+E_2(g_i^p+e_i^{(2)})\\
&+\omega\{S_re_i^{(r)}+E_r(u_i^p+e_i^{(r)})\}+A_{k_i}.
\end{align*}
Here $g_i^p$ and $u_i^p$ are point-feature norm bounds.
The assembly terms include the actual stored covariance relative to a two-pass point Gram, the latter Gram's enclosed inner-product rounding, and the final weighted sum, symmetrization, and ridge assembly.
The feature blocks are orthogonal coordinates of the combined feature representation, so each block contributes its own Gram perturbation.

With $A^*=K_{\omega,SS}+\tau I$, $A^*\succeq\tau I$.
For a stored system $\widetilde A$, stored cross vector $\widetilde k_i$, and any ordinary-precision dual approximation $w_i$, define
\[
t_i=\widetilde k_i-\widetilde A w_i.
\]
The query gain is bounded by
\[
C_i=\min\left\{
\frac{\sqrt{(g_i^p+e_i^{(2)})^2+\omega(u_i^p+e_i^{(r)})^2}}{2\sqrt\tau},\quad
\|w_i\|+\frac{\|t_i\|+d_{k_i}+d_K\|w_i\|}{\tau}
\right\}.
\]
This follows from the PSD ridge singular-value bound and the dual residual identity.
The same stored-solve perturbation formula as Equation~\eqref{eq:deploybound} gives simultaneous prediction radii.
All terms are evaluated in one consistent response unit and converted back to the original output unit.

\subsection{Complete-risk certificate}

Exact rational score ordering and Arb positive pivot sums enclose the complete risk of the saved predictions.
Theorem~\ref{thm:riskperturbation} propagates the simultaneous prediction radii to the exact finite-state predictor.
The calculation holds for the fixed observed labels, positive scales, and selected hyperparameters used in the run.
Independent Monte Carlo repetitions assess the sampling variation of the covariance estimator.

\IfFileExists{residual_certificate_results.tex}{The complete residual certificate covers Airfoil reference 4101 at $q=8,32$ and the heavy-tailed reference 4101 at $q=8$.
Each condition contains all 640 source functions and all 512 evaluation-query functions, at 8192 shared states and replicate 0.
Every fixed weight is certified. The independently selected weights are $0,1,1$ for the three conditions in that order.
The initial $10^{-9}$ coefficient contract is retained in the execution archive.
An error-radius driven refinement applies a $10^{-13}$ local contract to operators whose coefficient or mean radius exceeds $10^{-12}$.
\begin{center}
\small
\begin{tabular}{llrrrr}
\toprule
Condition & $\omega$ & Saved $D$ & Risk radius & Radius/$D$ (\%) & Exact finite-state $D$ interval\\
\midrule
Airfoil $q=8$ & $\mathbf{0}$ & $0.00017302$ & $7.080\times10^{-10}$ & $4.092\times10^{-4}$ & $[1.7301955\times10^{-4},1.7302097\times10^{-4}]$\\
Airfoil $q=8$ & $0.5$ & $0.00018011$ & $2.162\times10^{-9}$ & $1.200\times10^{-3}$ & $[1.8010833\times10^{-4},1.8011266\times10^{-4}]$\\
Airfoil $q=8$ & $1$ & $0.00018108$ & $3.461\times10^{-9}$ & $1.911\times10^{-3}$ & $[1.8107549\times10^{-4},1.8108242\times10^{-4}]$\\
Airfoil $q=32$ & $0$ & $0.00051204$ & $2.337\times10^{-9}$ & $4.563\times10^{-4}$ & $[5.1203571\times10^{-4},5.1204039\times10^{-4}]$\\
Airfoil $q=32$ & $0.5$ & $0.000512$ & $9.994\times10^{-9}$ & $1.952\times10^{-3}$ & $[5.1198979\times10^{-4},5.1200978\times10^{-4}]$\\
Airfoil $q=32$ & $\mathbf{1}$ & $0.00051646$ & $1.587\times10^{-8}$ & $3.072\times10^{-3}$ & $[5.1644719\times10^{-4},5.1647893\times10^{-4}]$\\
Heavy-tailed $q=8$ & $0$ & $0.00034715$ & $9.591\times10^{-11}$ & $2.763\times10^{-5}$ & $[3.4714611\times10^{-4},3.4714632\times10^{-4}]$\\
Heavy-tailed $q=8$ & $0.5$ & $0.00034651$ & $1.611\times10^{-10}$ & $4.648\times10^{-5}$ & $[3.4651298\times10^{-4},3.4651331\times10^{-4}]$\\
Heavy-tailed $q=8$ & $\mathbf{1}$ & $0.00034602$ & $2.257\times10^{-10}$ & $6.521\times10^{-5}$ & $[3.4601728\times10^{-4},3.4601774\times10^{-4}]$\\
\bottomrule
\end{tabular}
\end{center}
Bold weights identify the actual selected rules. Every risk interval conditions on the recorded source observations, scales, normal draws, and selected hyperparameters.
The reproduction notes give the independent high-precision verification procedures for these intervals.
}{}
\IfFileExists{certificate_execution_cost.tex}{\subsection{Numerical-certificate execution cost}
The following table records the full original coefficient work used by each frozen task pool, its error-radius refinement, and the sample-value enclosure work.
Each coefficient column sums the recorded elapsed time of the relevant worker tasks. The sample column sums the 1152 functionwise enclosure tasks at 8192 shared states.
\begin{center}
\small
\begin{tabular}{lrrrr}
\toprule
Condition & \shortstack{Original source\\balls (s)} & \shortstack{Original query\\balls (s)} & \shortstack{Refined\\balls (s)} & \shortstack{Sample\\enclosures (s)}\\
\midrule
Airfoil $q=8$ & 13.90 & 13.49 & 17.88 & 149.31\\
Airfoil $q=32$ & 1132.39 & 749.54 & 1670.65 & 1289.05\\
Heavy-tailed $q=8$ & 11.25 & 6.66 & 3.48 & 263.05\\
\bottomrule
\end{tabular}
\end{center}
Recorded parallel wall times are 432.74 s for the full sample-bound stage and 435.49 s for all 1073 coefficient refinements, using four workers.
The final propagation of the refined row bounds through all nine branches takes 100.18 s, and the independent high-precision audit takes 40.38 s.
These are recorded computations on the frozen numerical panel. Their task-time sums and parallel wall times represent their stated execution scopes.
}{}

\section{Residual covariance development and matched controls}
The fixed panel contains ten datasets, four fresh development references 4301--4304, both task sizes, and four independent Monte Carlo streams, giving 80 conditions and 320 algorithm runs.
Source and query--source covariance variability are estimated separately within each fixed condition using the ordinary sample variance across the four independent streams.
Averaging the six pairwise half-squared differences is algebraically equal to that four-stream sample variance.
The block statistic divides the sum of residual covariance entry variances by the corresponding sum for complete-operator MC at the same 8192 states.
\begin{figure}[ht]
\centering
\includegraphics[width=\textwidth]{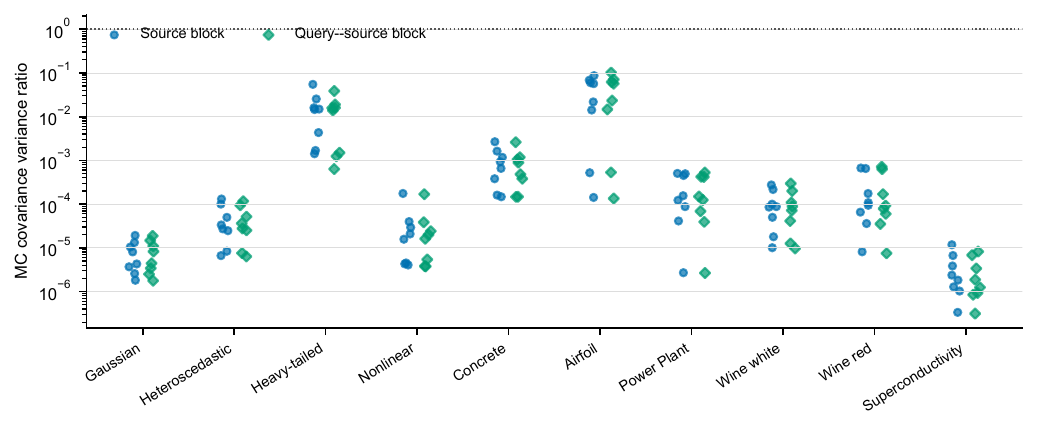}
\caption{Covariance sampling variation on every development condition. Each point represents one reference and task-size condition and is estimated from four independent MC streams.}
\end{figure}
\begin{center}
\small
\begin{tabular}{lrrr}
\toprule
Dataset & Residual/K2 & Residual/matched shrinkage & Residual/full MC 32768\\
\midrule
Gaussian & 1.000037 & 0.994009 & 0.994095\\
Heteroscedastic & 1.000256 & 0.993956 & 0.994851\\
Heavy-tailed & 0.990983 & 0.993062 & 0.991097\\
Nonlinear & 0.999986 & 0.995904 & 0.995882\\
Concrete & 0.999797 & 0.999130 & 0.985164\\
Airfoil & 0.979312 & 0.983680 & 0.991739\\
Power Plant & 1.000433 & 0.995021 & 0.994808\\
Wine white & 1.000517 & 0.994973 & 0.994059\\
Wine red & 0.999776 & 0.998229 & 0.993873\\
Superconductivity & 1.000083 & 1.000534 & 1.002004\\
\bottomrule
\end{tabular}
\end{center}

\IfFileExists{covariance_control_results.tex}{\section{Direct same-sample covariance control}
\label{app:covariancecontrol}
The registered development comparison reuses all ten datasets, references 4301--4304, both task sizes, and four independent Monte Carlo streams.
The stored 8192-state covariances supply the direct control-variate adaptation
\[ K^{CV}_\omega=K_2+\omega(\widehat C_H-\widehat C_{F_2}). \]
At a common state sample, its difference from the residual kernel is $\omega(\widehat C_{F_2,r}+\widehat C_{r,F_2})$.
The prediction adaptation retains the positive source eigenspace $V_+\Lambda_+V_+^\top$ and uses the consistent query readout
\[ p_i^{CV+}=\mu_i+k_i^\top V_+(\Lambda_++\tau I)^{-1}V_+^\top(Y_S-\mu_S). \]
Here $k_i=K^{CV}_{\omega,Si}$ is the source-to-query covariance column for the same weight.
Every family inherits the same degree-two physical ridge and scale and selects from the same three-point weight menu on the same selection tasks.
This is an application-specific covariance control motivated by polynomial-chaos control variates \citep{fox2021,yang2023}.
All 640 nonzero-weight source matrices have zero spectral correction on this development panel.
\begin{center}
\small
\begin{tabular}{lrrrr}
\toprule
Dataset & Residual/CV loss & 95\% interval & Source MC variance ratio & Query MC variance ratio\\
\midrule
Gaussian & 0.999539 & [0.998360,1.000625] & 0.002038 & 0.002059\\
Heteroscedastic & 1.000303 & [0.999453,1.001218] & 0.004846 & 0.004742\\
Heavy-tailed & 0.994094 & [0.990649,0.997831] & 0.08776 & 0.07559\\
Nonlinear & 0.999639 & [0.999184,1.000187] & 0.004204 & 0.003947\\
Concrete & 0.995984 & [0.991307,0.999829] & 0.0226 & 0.02043\\
Airfoil & 0.988485 & [0.977224,0.998644] & 0.1075 & 0.1153\\
Power Plant & 0.999928 & [0.998970,1.001124] & 0.01149 & 0.01081\\
Wine white & 1.000622 & [0.999190,1.002520] & 0.006565 & 0.006361\\
Wine red & 0.998146 & [0.996163,0.999841] & 0.007906 & 0.007773\\
Superconductivity & 0.999934 & [0.999415,1.000622] & 0.001172 & 0.0009664\\
\bottomrule
\end{tabular}
\end{center}
Loss ratios first average original-unit risks across references, task sizes, and streams within each dataset.
The paired bootstrap resamples references and then Monte Carlo streams, retaining the two task sizes together.
The covariance columns average the eight reference-and-task-size variance ratios within a dataset. Each variance estimate uses the four independent streams at a fixed condition.
\begin{figure}[ht]
\centering
\includegraphics[width=\textwidth]{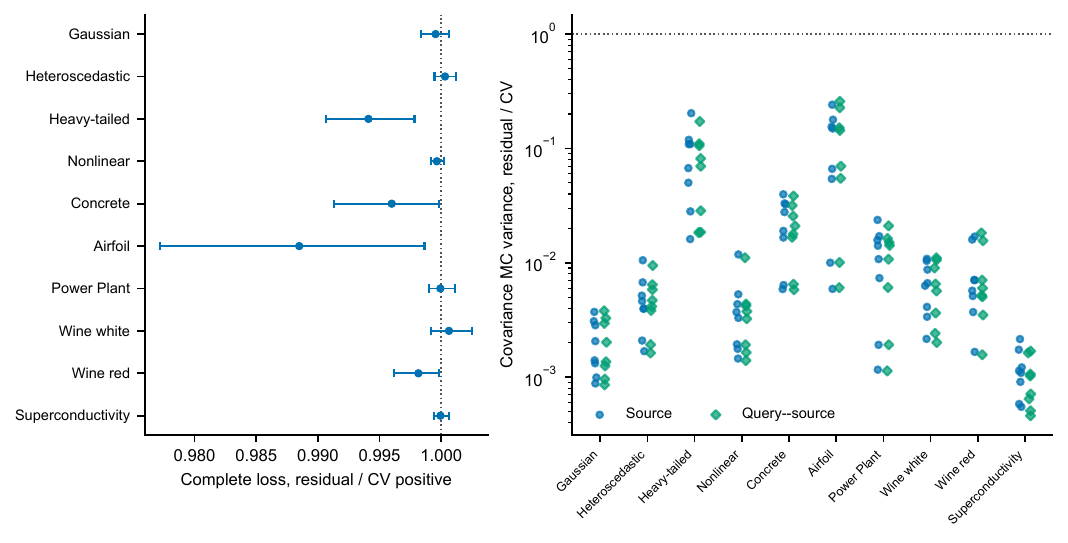}
\caption{Direct covariance-control comparison on the complete development panel. Left, complete-loss ratios with paired hierarchical bootstrap intervals. Right, every condition's ratio of covariance sampling variation, estimated separately for source and query--source blocks.}
\end{figure}
}{}
\IfFileExists{matched_family_cost_results.tex}{\section{Matched-family production time and memory}
\label{app:matchedcost}
Eight fixed development conditions cover Airfoil references 4101 and 4102 at both task sizes, and the heavy-tailed and Superconductivity reference 4101 at both task sizes.
Each family uses a fresh process and recomputes the common degree-two projection and kernel, the ten-ridge three-scale base selection, 8192 shared states, its covariance, and the three-weight selection.
The local host uses three single-native-thread workers and reserves logical CPUs 0--3. Family order rotates across conditions.
Production times sum the measured context, projection, kernel, base-learning, sampling-covariance, weight-learning, and complete-risk components.
Independent numerical checks and file I/O have their own recorded times. Peak process memory includes the interpreter and independent checking arrays.
\begingroup
\setlength{\tabcolsep}{3pt}
\begin{center}
\small
\begin{tabular}{llrrrrrr}
\toprule
Reference & $q$ & \shortstack{Residual\\(s)} & \shortstack{Full MC\\(s)} & \shortstack{CV positive\\(s)} & \shortstack{Residual\\(MiB)} & \shortstack{Full MC\\(MiB)} & \shortstack{CV positive\\(MiB)}\\
\midrule
Airfoil 4101 & 8 & 5.196 & 4.353 & 5.849 & 385.1 & 332.2 & 425.5\\
Airfoil 4101 & 32 & 11.980 & 9.799 & 13.224 & 417.6 & 365.4 & 458.0\\
Airfoil 4102 & 8 & 4.954 & 4.276 & 5.430 & 386.4 & 333.6 & 425.2\\
Airfoil 4102 & 32 & 11.986 & 9.518 & 12.540 & 417.1 & 365.5 & 459.6\\
Heavy-tailed 4101 & 8 & 7.541 & 6.249 & 8.094 & 583.8 & 530.9 & 593.7\\
Heavy-tailed 4101 & 32 & 19.456 & 15.241 & 19.602 & 621.2 & 569.4 & 631.4\\
Superconductivity 4101 & 8 & 16.585 & 15.345 & 16.904 & 1309.7 & 1256.7 & 1317.1\\
Superconductivity 4101 & 32 & 25.818 & 21.686 & 26.475 & 1368.3 & 1315.7 & 1377.4\\
\bottomrule
\end{tabular}
\end{center}
\endgroup
The complete-operator shrinkage family includes the same degree-two construction and base selection as the residual family.
Its sample covariance uses the complete-function values. Residual construction also evaluates the polynomial projection, while the direct control-variate family accumulates both complete and polynomial covariances.
The positive-source control additionally performs its spectral projection with a consistent query extension.
The execution records retain fit-and-forecast, selection, complete-risk, and numerical-audit costs separately for every family and condition.
}{}
\section{Source-observation learning curves}
\label{app:learningcurves}
Reference 4301 supplies all ten datasets and both task sizes.
The nested source budgets are 0, 80, 160, 320, and 640.
All budgets use the same 512-task selection pool and 512-task evaluation pool, with 256 calibration roles.
At zero source observations, the predictor is the analytic mean and its score exponent is chosen on the selection pool.
For each positive budget, the residual and matched complete-MC families inherit that budget's independently selected degree-two physical ridge and scale, then select their weight from the same three-point menu.
The matched complete-MC family uses 8192 states. The separate full-MC control uses 32768 states and independently selects between sample and matched means.
The experiment contributes 340 new dense complete-risk checks and retains the 480 checked risks from the positive-budget experiment.
\begin{center}
\small
\begin{tabular}{lrrrr}
\toprule
Dataset & $D_R(80)/D(0)$ & $D_R(160)/D(0)$ & $D_R(320)/D(0)$ & $D_R(640)/D(0)$\\
\midrule
Gaussian & 0.98837 & 0.96848 & 0.92859 & 0.87743\\
Heteroscedastic & 1.01043 & 1.00305 & 0.98714 & 0.93273\\
Heavy-tailed & 0.93795 & 0.91532 & 0.77790 & 0.72208\\
Nonlinear & 1.01092 & 1.01832 & 0.98772 & 0.92648\\
Concrete & 0.62952 & 0.57160 & 0.35655 & 0.22508\\
Airfoil & 1.01180 & 0.90869 & 0.81781 & 0.61915\\
Power Plant & 0.99482 & 0.96408 & 0.91523 & 0.83360\\
Wine white & 0.91753 & 0.86164 & 0.79528 & 0.66877\\
Wine red & 0.82683 & 0.73789 & 0.55814 & 0.42481\\
Superconductivity & 0.87229 & 0.83907 & 0.85518 & 0.76862\\
\bottomrule
\end{tabular}
\end{center}
Within a dataset, the original-unit risks are averaged across the two task sizes before taking each ratio.
The following table gives residual risk divided by the matched 8192-state full-MC shrinkage risk.
\begin{center}
\small
\begin{tabular}{lrrrr}
\toprule
Dataset & 80 sources & 160 sources & 320 sources & 640 sources\\
\midrule
Gaussian & 0.999987 & 0.999767 & 1.001250 & 0.997683\\
Heteroscedastic & 1.001319 & 0.994734 & 0.999947 & 0.987640\\
Heavy-tailed & 0.996586 & 0.997364 & 1.001165 & 0.985236\\
Nonlinear & 0.998694 & 0.999762 & 1.000101 & 0.986102\\
Concrete & 1.004015 & 1.003414 & 0.990289 & 0.986234\\
Airfoil & 0.999711 & 1.004008 & 1.005968 & 0.992660\\
Power Plant & 1.000009 & 1.000000 & 1.000000 & 0.990040\\
Wine white & 1.000138 & 1.001280 & 0.995281 & 0.999908\\
Wine red & 0.997954 & 0.987733 & 1.005969 & 1.000000\\
Superconductivity & 1.002890 & 0.996481 & 0.975734 & 1.000000\\
\bottomrule
\end{tabular}
\end{center}
\begin{figure}[ht]
\centering
\includegraphics[width=\textwidth]{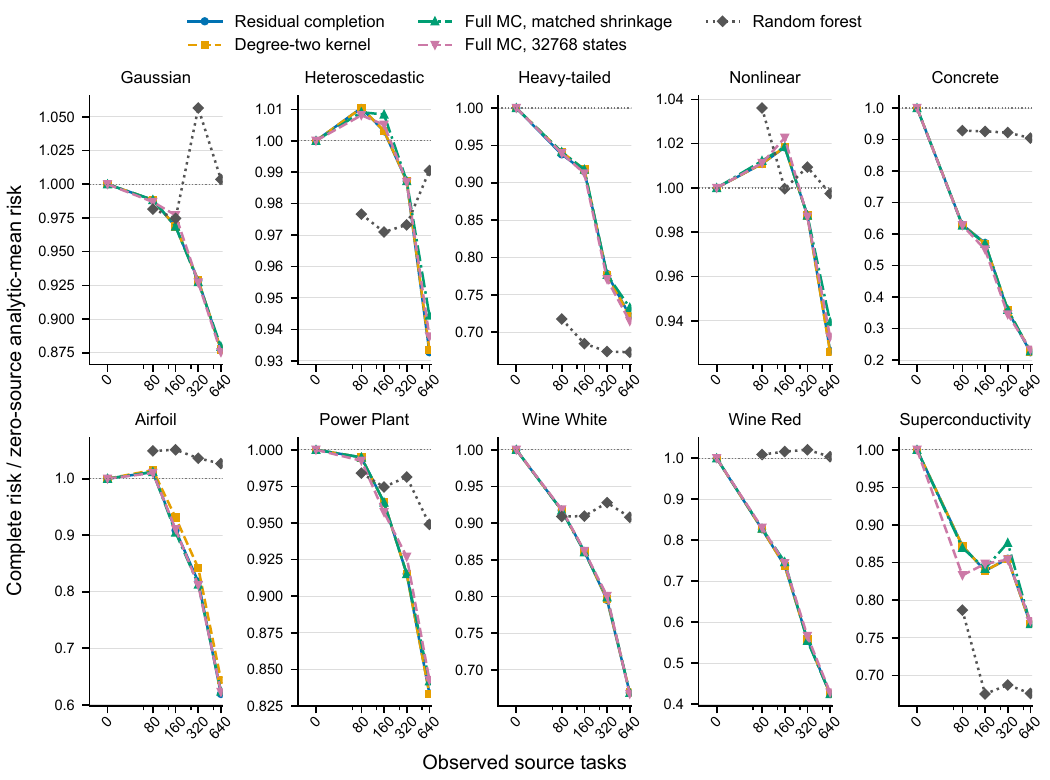}
\caption{Complete source-observation curves for all ten datasets at development reference 4301. Each panel uses that dataset's zero-source analytic-mean risk as its unit. The horizontal scale is linear from zero to 80 observations and logarithmic with base two above 80.}
\end{figure}
\begin{figure}[ht]
\centering
\includegraphics[width=\textwidth]{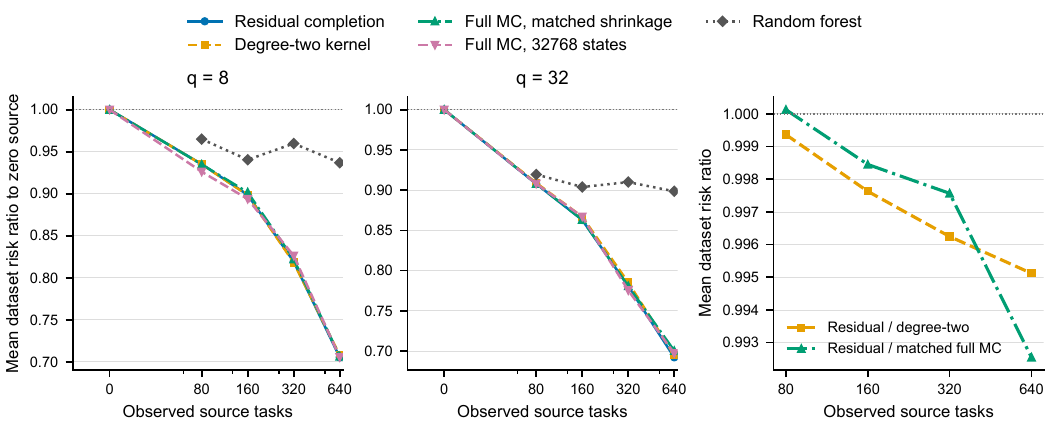}
\caption{Task-size strata and the residual covariance comparison on the same fixed development panel. The two learning panels average dataset ratios separately at each task size.}
\end{figure}

\end{document}